\documentclass[letterpaper, 10 pt, conference]{ieeeconf}  % Comment this line out if you need a4paper

\IEEEoverridecommandlockouts
\usepackage{graphicx}
\def\BibTeX{{\rm B\kern-.05em{\sc i\kern-.025em b}\kern-.08em
    T\kern-.1667em\lower.7ex\hbox{E}\kern-.125emX}}
\usepackage[font=small,labelfont=bf,tableposition=top]{caption}

\usepackage{blindtext}
\title{here title}

\makeatletter
\let\NAT@parse\undefined
\makeatother

\IEEEoverridecommandlockouts                              % This command is only needed if 
\usepackage[OT1]{fontenc}

\usepackage[square,numbers,sort&compress]{natbib}

\usepackage{bibentry}

\usepackage{amsmath}

\usepackage{amssymb}

\usepackage{mathtools}

\usepackage{cancel}

\usepackage{microtype}

\usepackage[plain]{algorithm}
\usepackage[noend]{algpseudocode}

\let\labelindent\relax
\usepackage[inline]{enumitem}

\usepackage{multicol}

\usepackage{booktabs}

\usepackage{subfig}

\usepackage[usenames,dvipsnames,table]{xcolor}

\usepackage[most]{tcolorbox}

\usepackage{nag}

\usepackage[disable]{todonotes}  % disable all todo notes.
\setuptodonotes{
  backgroundcolor=pink!30,
  textcolor=red!90!black,
  linecolor=red!90!black,
  bordercolor=red!90!black,
  caption={},  % to get enumerate/itemize to work (see https://tex.stackexchange.com/a/54068)
}

\usepackage{hyperref}

\definecolor{deeppink}{rgb}{1.0, 0.08, 0.58}
\hypersetup{
  colorlinks = true,  % colours link4.20s instead of ugly boxes
  urlcolor   = deeppink,  % colour for external hyperlinks
  linkcolor  = blue,  % colour of internal links
  citecolor  = red,  % colour of citations
  pdfview    = XYZ,  % page fit after navigation. Options: XYZ, Fit, FitH, FitV, FitB, FitR, FitBH, FitBV
}

\usepackage[nameinlink]{cleveref}  % nameinlink makes the whole name a link

\AtBeginEnvironment{appendices}{\crefalias{section}{appendix}}
\AtBeginEnvironment{appendices}{\crefalias{subsection}{appendix}}
\AtBeginEnvironment{appendices}{\crefalias{subsubsection}{appendix}}

\crefname{equation}{Eq.}{Eqs.}
\crefname{algorithm}{Alg.}{Algs.}
\crefname{figure}{Fig.}{Figs.}
\crefname{table}{Table}{Tables}
\crefname{tabular}{Table}{Tables}
\crefname{section}{Sec.}{Sections}
\crefname{appendix}{App.}{Appendices}

\creflabelformat{equation}{#2\textup{#1}#3}

\allowdisplaybreaks

\usepackage{hyperxmp}

\usepackage[acronym,toc]{glossaries}

\usepackage[export]{adjustbox}

\usepackage{tikz}

\usepackage[bottom,flushmargin]{footmisc}
\makeatletter  % blank footnote
\def\blfootnote{\gdef\@thefnmark{}\@footnotetext}
\makeatother

\usepackage[switch]{lineno}

\usepackage{dblfloatfix}

\usepackage{lipsum}

\tcbuselibrary{theorems}
\newtcbtheorem{boxtheorem}{\it Theorem}{colback=gray!30, colframe=black!60, title=\textit}{thm}

\usepackage{gensymb}

\usepackage{graphicx}

\usepackage{multirow}

\usepackage{makecell}

\newcolumntype{?}{!{\vrule width 1pt}}

\usepackage{pifont}

\usepackage{color, soul}

\newcommand{\seteqnspacing}[1]{
  \setlength{\belowdisplayskip}{#1} \setlength{\belowdisplayshortskip}{#1}
  \setlength{\abovedisplayskip}{#1} \setlength{\abovedisplayshortskip}{#1}
}

\newsavebox\CBox
\def\textBF#1{\sbox\CBox{#1}\resizebox{\wd\CBox}{\ht\CBox}{\textbf{#1}}}

\definecolor{darkred}{rgb}{0.55, 0.0, 0.0}

\newcommand{\smalltextbf}[1]{\textbf{\small #1}}

\newcommand{\ie}{\textit{i.e.}}
\newcommand{\fonestar}{F$_1^*$}

\newcommand{\classes}{{\mathbb{C}}}
\newcommand{\classesID}{{\classes_\mathrm{ID}}}
\newcommand{\classesOOD}{{\classes_\mathrm{OOD}}}

\newcommand{\iidsim}{\vcenter{\hbox{$\overset{\mathrm{i.i.d}}{\scalebox{1.5}[1]{$\sim$}}$}}}

\renewcommand{\mid}{\,\vert\,}
\newcommand{\N}{\mathcal{N}}
\newcommand{\train}{\mathrm{train}}

\newcommand{\edit}{\mathrm{edit}}
\newcommand{\inp}{\mathrm{input}}

\newcommand{\dataset}{\mathcal{D}_\train}

\newcommand{\weights}{{\boldsymbol\theta}\hspace{-0.05em}}
\newcommand{\bepsilon}{\boldsymbol\epsilon}

\newcommand{\dd}{\mathrm{d}}

\newcommand{\R}{\mathbb{R}}

\newcommand{\x}{\mathbf{x}}
\newcommand{\y}{\mathbf{y}}
\newcommand{\g}{\mathbf{g}}
\newcommand{\E}{\mathbb{E}}
\newcommand{\s}{\mathbf{s}}

\newcommand{\mufunc}[1]{\boldsymbol\mu_{#1}^\weights}
\newcommand{\epsfunc}[1]{{\boldsymbol\epsilon}_{#1}^\weights}

\newcommand{\rsim}{r_\mathrm{sim}}

\title{\LARGE \bf
\textit{Anomalies-by-Synthesis}: Anomaly Detection using\\Generative Diffusion Models for Off-Road Navigation
}

\newcommand{\website}{https://siddancha.github.io/anomalies-by-diffusion-synthesis}

\newcommand{\authornames}{
  Sunshine Jiang\textsuperscript{*}$^{1}$,
  Siddharth Ancha\textsuperscript{*}$^{1}$,
  Travis Manderson$^{1}$,
  Laura Brandt$^{1}$,\\
  Yilun Du$^{1}$,
  Philip R. Osteen$^{2}$ and
  Nicholas Roy$^{1}$% <-this % stops a space
}

\author{
  \authornames
  \\ \\ {\small Website: \href{\website}{\color{deeppink}\texttt{\website}}$^\dagger$}
  \vspace*{-0.05em}
  \thanks{
    \textsuperscript{*}Denotes equal contribution.
    $^{1}$MIT CSAIL, Cambridge, MA 02139, USA.
    $^{2}$DEVCOM Army Research Laboratory, Adelphi, MD 20783, USA.
  }%
  \thanks{
    Correspondence email: {\tt\small sancha@mit.edu}.
  }
  \thanks{
    $^\dagger$While this paper is fully self-contained, our project website contains (1) a 3-min overview video, (2) qualitative visualizations, (3) an appendix with detailed proofs, (4) Google Colab notebook, and (5) code, for easy access.
  }%
}

\let\oldtwocolumn\twocolumn
\renewcommand\twocolumn[1][]{
    \oldtwocolumn[{#1}{
        \vspace*{-0.2cm}
        \centerline
        {
          \includegraphics[trim=0 0 0 0,clip,width=\textwidth]{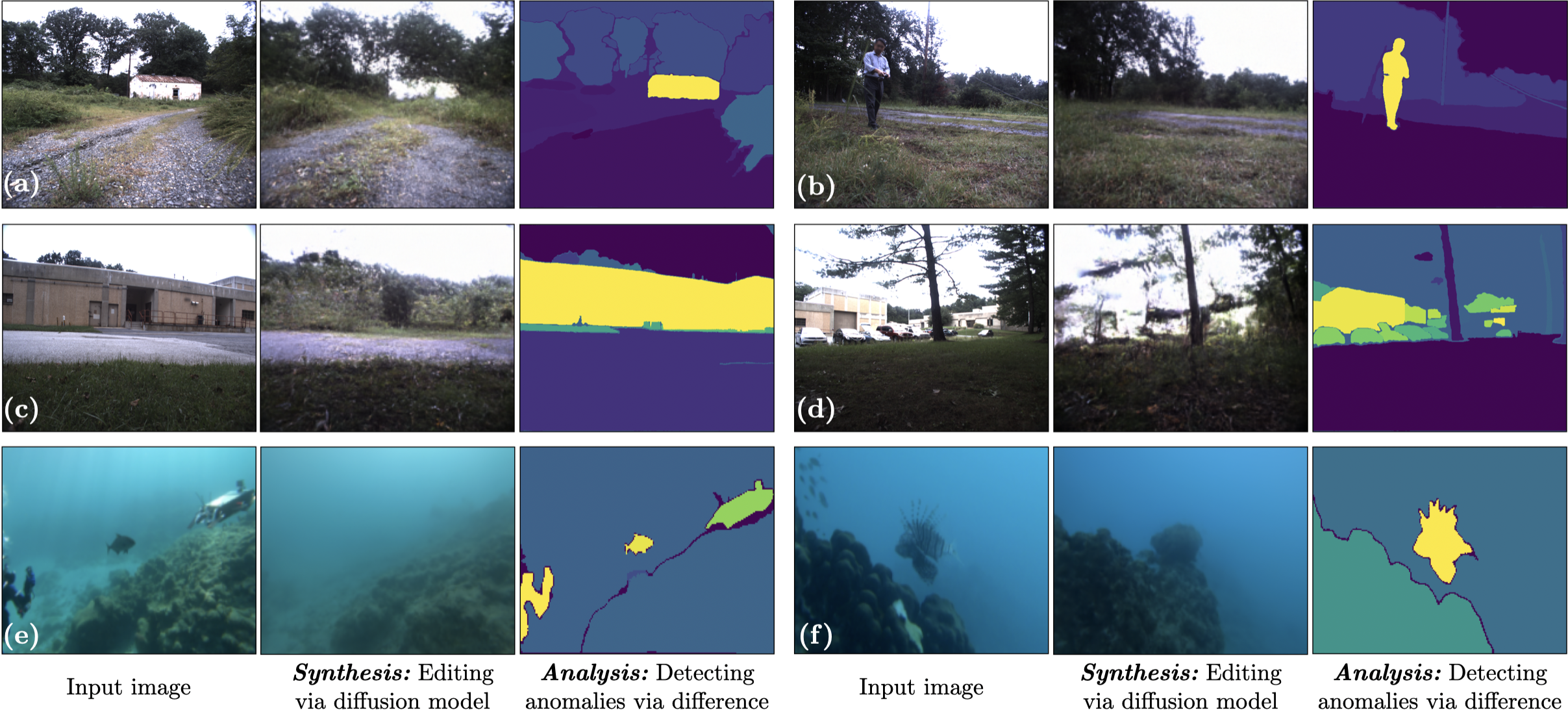}
        }
        \captionof{figure}
        {
          We present a method for anomaly detection in off-road images using an \textit{analysis by synthesis} approach.
          \textbf{\textit{Synthesis:}} A diffusion model trained on in-distribution data edits the input image to remove out-of-distribution segments. \textbf{\textit{Analysis:}} Anomaly detection is framed as extracting the difference between input and edited images.
          The training dataset contains off-road images with: (a-d) natural vegetation, ground and sky, but no buildings, humans or vehicles, (e-f) underwater scenes with mostly the oceanbed and coral.
          The diffusion model makes interesting edits, such as (a) blending buildings into sky, (b) removing people, (c) growing moss over buildings, (e) removing fish, robots and divers, and (f) morphing camouflaged fish into rocks. 
          Our method can detect small objects and multiple anomalies per image (d).
        }
        \label{fig:pull-figure}
        \vspace*{0.3cm}
    }]
}

\begin{document}
% \linenumbers

\maketitle
\thispagestyle{empty}
\pagestyle{empty}

%%%%%%%%%%%%%%%%%%%%%%%%%%%%%%%%%%%%%%%%%%%%%%%%%%%%%%%%%%%%%%%%%%%%%%%%%%%%%%%%

\noindent
\begin{abstract}
    In order to navigate safely and reliably in
    % urban and
    off-road and unstructured
    environments, robots must detect anomalies that are out-of-distribution (OOD) with respect to the training data.
    We present an \textit{analysis-by-synthesis} approach for pixel-wise anomaly detection without making any assumptions about the nature of OOD data.
    Given an input image, we use a generative diffusion model to \textit{synthesize} an edited image that removes anomalies while keeping the remaining image unchanged.
    % closer to the training distribution by removing anomalies.
    Then, we formulate anomaly detection as \textit{analyzing} which image segments were modified by the diffusion model.
    % We propose a novel approximate inference technique based on a principled analysis of guided diffusion that bootstraps the diffusion model to compute guidance gradients.
    We propose a novel inference approach for guided diffusion by analyzing the ideal guidance gradient and deriving a principled approximation that bootstraps the diffusion model to predict guidance gradients.
    Our editing technique is purely test-time that can be integrated into existing workflows without the need for retraining or fine-tuning.
    % allowing us to use pre-trained models.
    % Furthermore, our method is interpretable --- by synthesizing images that remove anomalies, we can inspect why the model believes certain regions are OOD.
    Finally, we use a combination of vision-language foundation models to compare pixels in a learned feature space and detect semantically meaningful edits, enabling accurate anomaly detection for off-road navigation.
    % Our diffusion-based analysis-by-synthesis approach produces accurate, interpretable anomaly detections for off-road navigation.
\end{abstract}

% {\keywords{Anomaly detection, Uncertainty estimation, Robot perception}}

%%%%%%%%%%%%%%%%%%%%%%%%%%%%%%%%%%%%%%%%%%%%%%%%%%%%%%%%%%%%%%%%%%%%%%%%%%%%%%%%

\begin{figure*}[t!]
    \centerline{
    \includegraphics[width=0.97\textwidth]{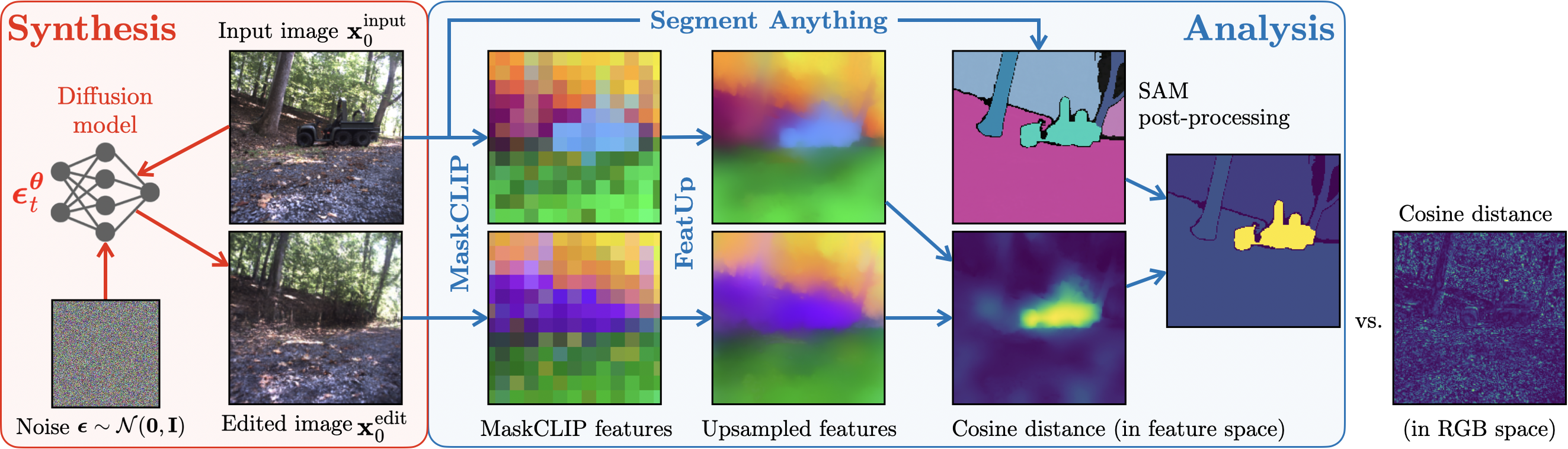}
    }
    \caption{
        \small
        Our proposed pipeline for pixel-wise anomaly detection.
        \textit{Left to right:} In the synthesis step, a trained diffusion model edits a given input image to remove anomaly segments without modifying other parts of the image.
        In this case, the model blends the OOD vehicle into dirt in the background.
        The analysis step extracts anomalies by comparing the pair of images in the CLIP~\cite{radford2021clip} feature space.
        First, MaskCLIP~\cite{dong2023maskclip} computes low-resolution CLIP features for each image, which are upsampled using FeatUp~\cite{fu2024featup}.
        In this figure, features are visualized via a t-SNE projection to three dimensions.
        Cosine distances between pixel features in the two images produce a raw anomaly map that highlights anomaly objects.
        In contrast, comparing the images directly in RGB space (extreme right) is noisy and unable to isolate OOD segments.
        Finally, SAM~\cite{kirillov2023segment} processes the input image to generate segments; these are used to refine and clean the anomaly map.
        \vspace*{-12pt}
    }
    \label{fig:pipeline}
\end{figure*}

\section{Introduction}
\label{sec:introduction}

The need for reliable autonomous robotic navigation is increasing in unstructured, off-road environments like planetary exploration \cite{bares1989ambler,massari2004autonomous}, forests \cite{frey2022locomotion}, deserts~\cite{mengterrainnet,shaban2022semantic} and underwater~\cite{eustice2008visually,manderson2020rss}.
However, the perception systems required for off-road navigation (e.g., semantic segmentation~\cite{valada2017deep,valada2017adapnet,guan2022ga}) are often trained on relatively small datasets~\cite{RUGD,RELLIS} and deployed in environments where images often contain ``out-of-distribution'' (OOD) \textit{anomalies} not well represented in the training data.
% For example, training data might not contain any images of humans, but humans may be present during deployment. 
Furthermore, lighting conditions, weather, and terrain can vary significantly between training and test environments.
% Robots operating in such real world environments often encounter \textit{anomalies} in images during deployment that are ``out of distribution" (OOD) with respect to the training data.
To navigate safely and reliably in unfamiliar environments, robots must identify such \textit{anomalies} in order to anticipate potential perception failures.
% and query human operators for help.
% Furthermore, it is helpful to localize anomalies at the pixel or segment level rather than at the image level.
% For example, if the anomalous segment is not in the robot's path, the robot can continue to navigate safely without human intervention.
The aim of this work is to detect anomalous segments (if any) in a given RGB image that are out-of-distribution from the training data.

Prior work on anomaly detection has focused on \textit{discriminative} models that directly map pixels to anomaly scores. These models classify features at the pixel- or segment- level as in-distribution or as anomalies~\cite{ancha2024icra,ulmer2021survey,charpentier2020posterior,charpentier2022natural,grcic2024dense,liu2023residual,hendrycks2016baseline,lakshminarayanan2017simple,lee2018simple,gudovskiy2023concurrent,tian2022pixel,bevandic2022dense,chan2021entropy,nayal2023rba,ackermann2023maskomaly}.
On the other hand, recent advances in generative AI have led to the development of diffusion models~\cite{ho2020denoising,diffusion_survey} that can model complex distributions and generate high-quality realistic images~\cite{dhariwal2021diffusion,rombach2022high}.
In this work, we present an \textit{\textbf{analysis-by-synthesis}}~\cite{bever2010analysis,yuille2006vision} approach to anomaly detection (hence the title ``anomalies-by-synthesis'').
We use a \textit{generative} diffusion model to ``edit'' the input image and remove anomalies while keeping the remaining input unchanged.
The new, \textit{\textbf{synthesized}} image represents what the scene would have looked like had it not contained any anomalies.
Then, we frame the problem of anomaly detection as \textit{\textbf{analyzing}} which parts of the image were edited by the generative model
% to bring the image in-distribution.
to remove anomalies.
In other words, the difference between the input and synthesized images produces anomaly detections.

Our method has several advantages. First, we
\textBF{do} \textBF{not} \textBF{make} \textBF{any} \textBF{assumptions} \textBF{about} \textBF{the} \textBF{nature} \textBF{of} \textBF{anomalies}
% or OOD examples that
the model can expect at test time.
% Additionally
Our approach \textBF{does not require} \textBF{any OOD data during training}.
% The \textit{analysis-by-synthesis} formulation is \textBF{more interpretable} than directly predicting anomaly masks.
% Synthesizing images that remove anomalies
% % we can visualize \textit{why} the diffusion model considers certain segments to be OOD.
% can provide insights about properties of the training data.
% For example, if the diffusion model removes people from the image, it hints that humans were absent in the training set.
% If the model lightens parts of the image, it highlights a gap in lighting conditions between training and test environments.
% % Analysis-by-synthesis offers substantially more \textit{interpretability} compared to directly predicting anomaly masks.
% Such insights could guide practitioners on how best to collect more training data and improve their model performance.
We use diffusion guidance \cite{sohl2015deep,song2020score,dhariwal2021diffusion,du2023reduce} to edit the input image. This is a post-hoc test-time procedure that \textBF{does not require re-training or} \textBF{finetuning the diffusion model}.
% or modifying its training procedure.
The diffusion model is only assumed to be trained on the standard objective of fitting to the training data.
% it does not need to be specialized to edit input images at test-time.
Therefore, our method can be applied to pre-trained or off-the-shelf diffusion models whose architectures and hyperparameters are carefully tuned for high-quality image synthesis.
% Therefore, our method can be directly integrated into existing diffusion workflows whose architectures and hyperparameters are carefully tuned for high-quality image synthesis.
% without needing to re-train or finetune the diffusion model.

We propose a \textBF{novel diffusion guidance approach} to ensure that the synthesized image is both similar to the input image and has a high probability under the training distribution.
We \textBF{theoretically analyze the ideal guidance gradient}, which is intractable to compute.
We then \textBF{propose a principled and} \textBF{tractable approximation} that, unlike prior methods, reuses the learned diffusion score function to compute the guidance gradient.
Our diffusion guidance approach modifies pixels corresponding to anomaly regions while keeping the in-distribution parts of the image as close to the input as possible.

Unfortunately, extracting anomaly segments from the edited image
% synthesized by diffusion models
is not as straightforward as computing pixel-wise intensity differences with the input image.
The diffusion model tends to slightly alter pixels throughout the entire image, including regions that are in-distribution.
While the differences are subtle to the human eye, pixel-wise intensity differences are sensitive to even minor changes between images.
This sensitivity produces extremely noisy anomaly masks riddled with false positives.
We mitigate this issue using a \textBF{combination of foundation vision models}: MaskCLIP~\cite{dong2023maskclip}, FeatUp~\cite{fu2024featup} and SAM~\cite{kirillov2023segment}.
By comparing images in CLIP~\cite{radford2021clip} feature space, we are able to \textBF{detect semantically meaningful edits while being robust to} \textBF{subtle and inconsequential pixel intensity changes}.

We quantitatively validate the effectiveness of our approach on two public off-road land navigation datasets: RUGD~\cite{RUGD} and RELLIS~\cite{RELLIS}, and qualitatively evaluate on an underwater navigation dataset~\cite{manderson2020rss}.

\section{Problem formulation: Anomaly detection as post-hoc analysis-by-synthesis}
\label{sec:problem-formulation}

\textbf{Pixel-wise anomaly detection:} We consider the problem of pixel-wise anomaly detection in RGB images.
We denote an RGB image of dimensions $H$$\times$$W$ by $\x_0 \in [0, 1]^{3HW}$, with a zero-subscript\footnote{The zero-subscript in $\x_0$ is introduced for consistency with notation used for diffusion models~\cite{ho2020denoising} (see \cref{sec:diffusion-background}). It corresponds to the noise level \hbox{$t=0$} implying that no noise has been added to the image. We slightly modify conventional notation to consistently specify timesteps in the subscript and model parameters in the superscript when applicable e.g. $\x_t, p^\weights, \epsfunc{t}$, $\mufunc{t}$.}.
% The image contains $H\times W$ pixels indexed by the set $\I$ where each pixel $i\in\I$ contains RGB values $\x_{0, i} \in [0, 1]^3$.
In our problem setting, we are given a set of $N$ training images $\dataset = \{\x_0^{(n)}\}_{n=1}^N ~\iidsim~ q(\x_0)$
% where each image is
assumed to be sampled \textit{i.i.d.} from a training distribution\footnote{We denote the training distribution by $q(\cdot)$ following \citet{ho2020denoising}.} $q(\x_0)$.
We refer to $q$ as the ``training distribution" or the ``in-distribution".
% For example, $q$ in our experiments contains off-road images with natural vegetation (trees, bushes, grass), ground (mud, gravel) and sky, but no humans, buildings or vehicles. 
At test time, we are given an input image $\x_0^\inp$.
A subset of its pixels belong to objects that are \textit{out-of-distribution} (OOD).
%  such as people, vehicles or buildings.
The task is to detect these anomalous pixels in the image.
During training, the anomaly detector is not allowed to make any assumptions about the nature of OOD examples it can expect at test time.
Anomalies are \textit{implicitly} defined by how unlikely the respective object segments are under the training distribution $q(\x_0)$.

\textbf{Analysis-by-synthesis formulation:}
% We formulate the problem of pixel-wise anomaly detection using the ``analysis by synthesis" framework in three stages.
We formulate the problem of pixel-wise anomaly detection in three stages.

\textbf{(i) \textit{Training:}} We train a whole-image generative model $p^\weights(\x_0)$, parameterized by weights $\weights \in \R^{W}$, to fit the training set $\dataset$ and learn $q(\x_0)$.
Learning $p^\weights(\x_0) \approx q(\x_0)$ is the only objective in this stage; training is not specialized for anomaly detection.
This de-coupling of the problem of \textit{learning} $p^\weights \approx q$ from the task of \textit{inferring} anomaly segments given a learned model $p^\weights$ allows us to repurpose pre-trained generative models for anomaly detection.

\textbf{(ii) \textit{Synthesis:}} Given an input image $\x_0^\inp$ at test time, we wish to detect anomaly segments by synthesizing an edited image using the learned model $p^\weights \approx q$.
We assume a similarity metric $\rsim: \R^{3HW} \times \R^{3HW} \rightarrow \R_{\geq 0}$ between two images that quantifies how similar they are.
In this work, we use the Gaussian kernel: \hbox{$\rsim(\x_0', \x_0'') = \exp(-\lambda \|\x_0' - \x_0''\|_2^2)$}, although any non-negative similarity metric can be used.
We define the editing process, i.e., \textit{synthesis}, as sampling from the following distribution:
{
\seteqnspacing{6pt}
\begin{align}
    \x_0^\edit \sim q(\x_0^\edit \mid \x_0^\inp) \propto \hspace{0.1em} \underbrace{\vphantom{\big|}q(\x_0^\edit)}_{\hspace{-11em}\text{likelihood under training distribution}\hspace{-3.5em}} \hspace{-0.1em}\underbrace{\vphantom{\big|}\rsim(\x_0^\edit, \x_0^\inp)}_{\hspace{0.9em}\text{similarity to input image}}
\label{eqn:synthesis}
\end{align}
}
The distribution in \cref{eqn:synthesis} is a product of two terms. The first term $q(\cdot)$, which is approximated by the learned generative model $p^\weights(\cdot)$, ensures that the edited image $\x_0^\edit$ is likely under the training distribution.
The second term ensures that $\x_0^\edit$ is similar to the input image.
% This editing procedure effectively synthesizes a new image that removes anomalies from $\x_0^\inp$ but keeps pixels in the in-distribution regions as close as possible to the input image.
Sampling from this conditional distribution synthesizes a new image that effectively ``edits'' $\x_0^\inp$ by removing anomalies but keeping pixels in the in-distribution regions as close as possible to the input image.
% In other words, we formulate detecting anomalies in an input image (\textit{analysis}) as editing its OOD segments (\textit{synthesis}).
% $\x_0^\edit$ is the ``counterfactual'' of what the input image might have looked like, had it not contained any anomalies.
$\x_0^\edit$ can also be interpreted as a ``projection'' of $\x_0^\inp$ onto the manifold of the training distribution $p^\weights(\x_0) \approx q(\x_0)$.
Sampling from \cref{eqn:synthesis} is a challenging inference problem: $q(\x_0)$ is a complex multimodal distribution over natural images, and multiplying by $\rsim(\cdot)$ makes the distribution unnormalized with an intractable normalizing constant.
In this work, we develop a diffusion guidance approach~\cite{sohl2015deep,song2020score,dhariwal2021diffusion} to sample from this distribution.
The synthesis procedure is ``post-hoc" in the sense that the generative model is not explicitly trained to perform the editing task; editing is result of an inference procedure performed at test-time.

\textbf{(iii) \textit{Analysis:}} We formulate detecting anomalies in $\x_0^\inp$ as computing the difference between $\x_0^\inp$ and $\x_0^\edit$.
As described in \cref{sec:introduction}, RGB intensities in $[0, 1]^3$ may not be a suitable space for this comparison.
We define the \textit{analysis} task as finding an appropriate $C$-dimensional feature space $f: [0, 1]^{H \times W \times 3} \rightarrow \R^{H \times W \times C}$ so that the anomaly score for the $i$-th pixel can be computed as $\|f(\x_0^\edit)_i - f(\x_0^\inp)_i \|$.
We use the CLIP~\cite{radford2021clip} feature space for comparison, via a combination of MaskCLIP~\cite{dong2023maskclip}, FeatUp~\cite{fu2024featup} and SAM~\cite{kirillov2023segment}.

\begin{figure}[t!]
    \centering{
        \includegraphics[width=0.43\textwidth]{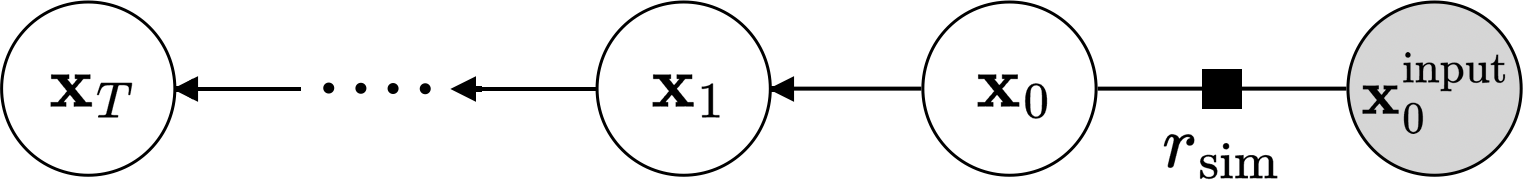}
    }
    \caption{
        \fontsize{9pt}{9pt}\selectfont
        Probabilistic graphical model for the \textit{conditional} forward diffusion process.
        The target variable we wish to sample is $\x_0$.
        Directed edges correspond to the standard forward diffusion process.
        The unnormalized factor $\rsim(\x_0, \x_0^\inp)$ conditions $\x_0$ to be similar to the (fixed) input image.
    }
    \label{fig:pgm}
\end{figure}

\begin{algorithm*}[t!]
    \newcommand{\format}[1]{{\fontsize{10pt}{10pt}\color{red}#1}}
    \centerline{
    \includegraphics[width=0.95\textwidth]{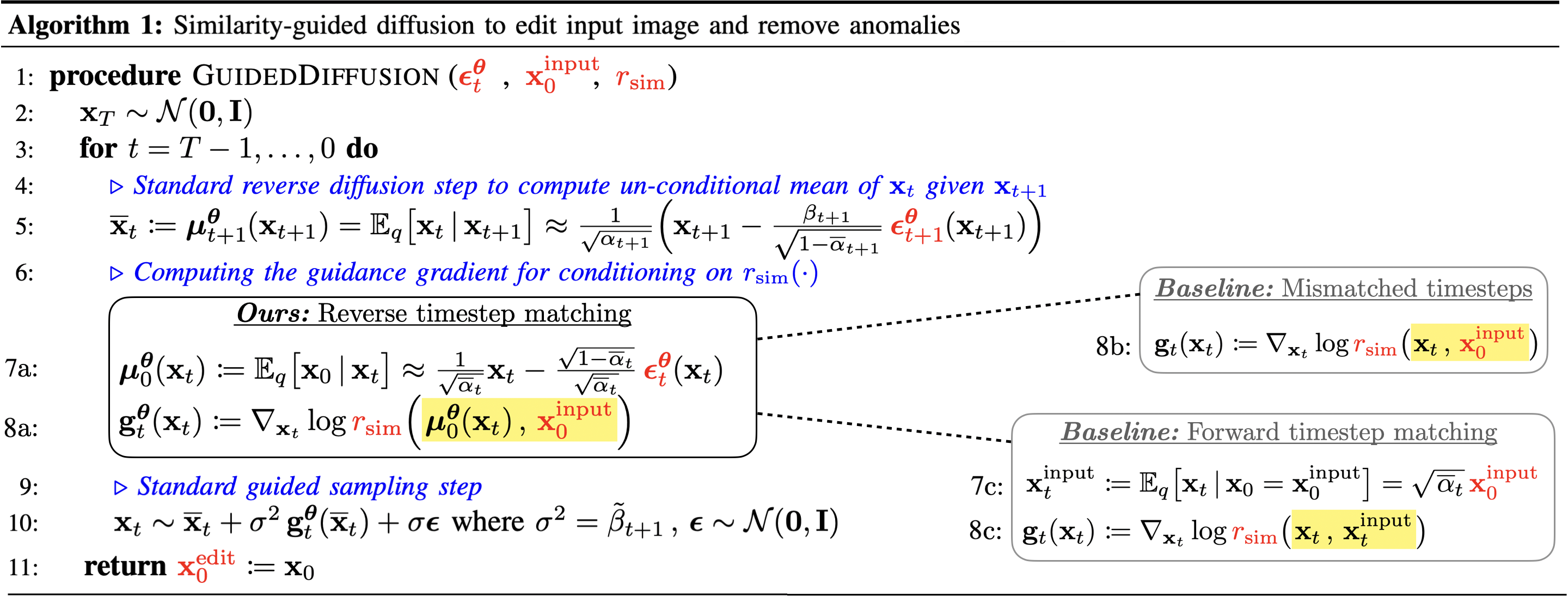}
    }
    \caption{
        \small
        \textit{Inputs:} (1) \format{$\epsfunc{t}$}: denoising diffusion model fit to training image distribution $q(\x_0)$.
        (2) \format{$\x_0^\inp$}: input image at test-time potentially containing anomaly segments.
        (3) \format{$\rsim$}: similarity metric between two images. We use $\rsim(\x, \y) = \exp\left( -\lambda \|\x - \y\|_2^2 \right)$.
        \textit{Output:} \format{$\x_0^\edit$}: edited version of $\x_0^\inp$ sampled from $q(\x_0)\,\rsim(\x_0, \x_0^\inp)$ that removes anomalies.
        % Our algorithm modifies an input image $\x_0^\inp$ with potentially OOD segments
        % Our algorithm modifies $\x_0^\inp$ to be similar to the training distribution $q(\x_0)$ using classifier-based guidance~\cite{sohl2015deep,dhariwal2021diffusion}.
        While standard reverse diffusion (lines 5, 10) is designed to sample from $q(\x_0)$, the image similarity metric $\rsim(\x_0, \x_0^\inp)$ guides the generated image $\x_0$ to be close to $\x_0^\inp$ using guided diffusion~\cite{sohl2015deep,dhariwal2021diffusion}.
        This is achieved by including the log-gradient of $\rsim$ (line 8a).
        %  that results in samples from the modified distribution $\x_0 \sim q(\x_0)\,\rsim(\x_0, \x_0^\inp)$.
        A key question during intermediate timesteps $t$
        % of the reverse diffusion process
        is what pair of images should the similarity metric compare?
        A na\"ive approach~\cite{sohl2015deep} (line 8b) simply uses the intermediate sample $\x_t$ and $\x_0^\inp$; they however correspond to different noise levels and should not be compared directly.
        Another baseline attempts to bring $\x_0^\inp$ to timestep $t$ by expected forward diffusion (lines 7c, 8c).
        Instead, we \textit{propose using the trained diffusion model} to estimate the expected noise-free version $\mufunc{0}(\x_t)$ of $\x_t$ to compare with the input image $\x_0^\inp$ at $t=0$.
        While this requires a backward pass through the diffusion model (line 8a),
        we show that our approach is mathematically principled and empirically leads to better performance.
        % {\color{red}\textbf{TODO:} (1) Add line numbers to baseline blocks.}
        \vspace{-1.5em}
    }
    \label{alg:conditional-diffusion}
\end{algorithm*}

\setlength{\textfloatsep}{10pt} % \input{graphics/algos/algo-latex.tex}
\section{Background on diffusion models}
\label{sec:diffusion-background}

% \noindent
Diffusion models~\cite{sohl2015deep,ho2020denoising,song2020score} are recently developed, state-of-the-art generative models effective at learning and sampling from complex, multi-modal image distributions.
% They have shown to be extremely effective in learning complex, multi-modal image distributions.
They have been primarily used in the computer vision community for photorealistic image generation and creative applications~\cite{ramesh2021zero, rombach2022high}, and in the robot learning community to learn distributions over action trajectories~\cite{chi2023diffusion, team2024octo}.
In this work, we use diffusion models for anomaly detection in images.
Recall that we denote (noiseless) image vectors by $\x_0 \in [0, 1]^{3HW}$.
% , and the training distribution as $p(\x_0) = q(\x_0)$.
We train a denoising diffusion probabilistic model (DDPM)~\cite{ho2020denoising} $p^\weights(\x_0)$ to fit the training distribution $q(\x_0)$ and sample from it.
To do so, DDPM introduces $T$ latent variables $\x_1, \dots, \x_T$ by the \textit{forward diffusion process} \hbox{$q(\x_t \mid \x_{t-1}) = \N\left( \x_t\,;\,\sqrt{1-\beta_t}\,\x_{t-1}, \beta_t \bf{I} \right)$}.
At each timestep $t$, forward diffusion progressively adds a small amount of \textit{i.i.d.} Gaussian noise with variance \hbox{$\beta_t > 0$} to $\x_{t-1}$.
With an appropriate noise schedule $\beta_{1:T}$, the final latent variable $\x_T$ is approximately normally distributed i.e. $\x_T \sim \N(\bf{0}, \bf{I})$.

% Since $q(\x_{1:T} \mid \x_0)$ is Gaussian.
% The conditional distribution $q(\x_t \mid \x_0)$ is also Gaussian and can be computed in closed form as $q(\x_t \mid \x_0) = \N(\x_t \,;\, \sqrt{\overline{\alpha}_t} \x_0, (1-\overline{\alpha}_t)\bf{I})$ where $\overline{\alpha}_t = \prod_{s=1}^t (1 - \beta_t)$ \cite{sohl2015deep,ho2020denoising}.
Because forward diffusion is Gaussian, latent variables from intermediate timesteps $\x_t$ can be directly sampled from \hbox{$\x_0$ as $\x_t(\x_0, \bepsilon) = \sqrt{\overline{\alpha}_t}\,\x_0 + \sqrt{1-\overline{\alpha}_t}\,{\bepsilon}$} where \hbox{$\overline{\alpha}_t = \prod_{s=1}^t (1 - \beta_t)$} \cite{sohl2015deep,ho2020denoising} and \hbox{${\bepsilon} \sim \N(\bf{0}, \bf{1})$} is a noise vector sampled from the standard normal distribution.
% is noise sampled from the standard Normal distribution.
% Given $\x_0 \sim q(\x_0)$ from the training set,
DDPM trains a neural network $\epsfunc{t}(\x_t)$ with weights $\weights$ to predict $\bepsilon$ that generated $\x_t$.
$\epsfunc{t}$ is trained to minimize the loss \hbox{$\frac{1}{T}\sum_{t=1}^T \E_{\x_0 \sim q(\x_0), \bepsilon \sim \N(\mathbf{0}, \mathbf{I})}\|\bepsilon - \epsfunc{t}(\x_t(\x_0, \bepsilon))\|_2^2$}.
% i.e. it minimizes the loss $\|\bepsilon - \epsfunc{t}(\x_t)\|_2^2$.
%  by regressing $\x_t$ to the standard Gaussian noise $\epsilon \sim \N(\bf{0}, \bf{1})$ produced $\x_t$ from $\x_0$ i.e. $\x_t = \sqrt{\overline{\alpha}}\x_0 + \sqrt{1-\overline{\alpha}_t}\,\epsilon$.
This network allows DDPM to perform \textit{reverse diffusion}.
First, $\x_T \sim p^\weights(\x_T) = \N(\mathbf{0}, \mathbf{I})$ is sampled from the standard normal.
Then, $\x_T$ is progressively denoised to $\x_0$. In the limit of small $\beta_t$, the reversal of forward diffusion $q(\x_t \mid \x_{t+1})$ becomes Gaussian~\cite{sohl2015deep} and
% so a neural network parametrized by $\theta$ is trained to predict the Gaussian parameter $\mufunc(\x_t, t)$ of reverse diffusion: $p_\theta \left( \x_{t-1} \mid \x_t \right) = \N\left( \x_{t-1} \,;\, \mufunc(\x_t, t), \tilde{\beta_t} \right) \approx q(\x_{t-1} \mid \x_t)$. See \citet{ho2020denoising} for more details.
can be approximated as $q(\x_t \mid \x_{t+1}) \approx p^\weights \left( \x_t \mid \x_{t+1} \right) = \N\big( \x_t \,;\, \mufunc{t+1}(\x_{t+1}), \tilde{\beta}_{t+1} \big)$, where $\mufunc{t+1}$ is a reparametrization of $\epsfunc{t+1}$ (see line 5 of \cref{alg:conditional-diffusion}).
Finally, the reverse diffusion process produces a sample from $p^\weights(\x_0) \approx q(\x_0)$.
We refer to \citet{ho2020denoising} for more details on DDPM.

% Diffusion models are a latent variable model where $\{x_i\}_{i=1}^T$ the latent variables. The joint distribution $p(x_{0:T})$ can be described by the \textit{forward (or noising) process}, $p(x_{0:T}) = p(x_0) \prod_{i=1}^T p(x_i | x_{i-1})$ where $p(x_i | x_{i-1}) = \sqrt{1-\beta_t}x_{i-1} + \beta_t \epsilon$ and $\epsilon \sim N(0, I)$. With an appropriate schedule of $\beta_t$, $x_T \sim N(0, I)$. The joint distribution can also be described by the \textit{reverse (or denoising) process}, $p(x_{0:T}) = p(x_T)\prod_{i=0}^{T-1}p(x_i | x_{i+1})$. \erick{I eventually want to get to the point that for the forward process, we can compute $p(x_t|x_0)$ can be sampled in closed form. This is often taken advantage of to perform in painting by replacing the observed portion with a noised version of the observation. I may have made a mistake, feel free to make any corrections.}

% \section{Guided diffusion for conditional sampling}
\section{Guidance gradients for conditional diffusion}
% \input{graphics/pgm.tex}

% \noindent {\small \textbf{Conditional guidance gradient:}}
Given an input image $\x_0^\inp$, we are interested in sampling from {$q(\x_0 \mid \x_0^\inp) \propto q(\x_0)\,\rsim(\x_0, \x_0^\inp)$} as noted in \cref{eqn:synthesis}.
To do so, we first define a \textit{conditional forward diffusion process}~\cite{dhariwal2021diffusion} as shown in \cref{fig:pgm}.
It can be denoted by {$q(\x_{0:T} \mid \x_0^\inp) = q(\x_0 \mid \x_0^\inp)\,q(\x_{1:T}\mid\x_0)$}, where $\x_0$ is hypothetically sampled from our desired conditional distribution $q(\x_0 \mid \x_0^\inp)$,
% in \cref{eqn:synthesis},
% $p'(\x_0) \propto p(\x_0)\,r(\x_0)$ obtained by weighting the original distribution $p(\x_0)$ with an un-normalized energy function $r(\x_0)$,
followed by the standard forward diffusion process $q(\x_{1:T} \mid \x_0)$.
%  as previously described.
% The learned diffusion model
% $p^\weights(\x_{t-1} \mid \x_{t})$
% $\epsfunc{t}(\x_t)$
% only reverses the \textit{vanilla} forward diffusion process.
A compelling property of diffusion models is that $\epsfunc{t}$ can be used to reverse a \textit{conditional} forward diffusion processes at test-time without having to re-train $\epsfunc{t}$.
Specifically, \textit{classifier guidance}~\cite{sohl2015deep,song2020score,dhariwal2021diffusion,du2023reduce} can be used in the special case where one wishes to sample from $q(\x_0 \mid y) \propto q(\x_0)\,q(y \mid \x_0)$ \ie, the conditioning factor takes the form of a classifier $q(y \mid \x_0)$,
% trained on intermediate latent states $\x_t$,
% \textit{classifier guidance}~\cite{sohl2015deep,song2020score,dhariwal2021diffusion}
% can sample from the conditional reverse diffusion process by
Classifier guidance adds a gradient step $\g_t(\x_t) = \nabla_{\x_t} \log p(y \mid \x_t)$ at each reverse diffusion step, shown in line 8 (a, b, c) of \cref{alg:conditional-diffusion}.
Intuitively, the guidance gradient encourages the conditioning factor of the resulting $\x_t$ to increase, while balancing with the original objective of sampling from $q(\x_0)$.

{\small \textbf{Mismatched-timesteps baseline:}} In our case, the conditioning factor is not a classifier but is of the form $\rsim(\x_0, \x^\inp)$.
A na\"ive application of classifier guidance, as used in \citet{sohl2015deep}, would be to choose the guidance gradient as $\g_t(\x_t) = \nabla_{\x_t} \log \rsim(\x_t, \x_0^\inp)$; see line 8b of \cref{alg:conditional-diffusion}.
However, this formulation compares a noiseless image $\x_0^\inp$ with a noisy image $\x_t$ corresponding to the noise level at timestep $t$.
The mismatch between the timesteps of the two images being compared leads to poor performance.
% We treat this method as a baseline called ``mismatched timesteps''.

{\small \textbf{Forward timestep matching baseline:}} A heuristic approach to match the timesteps of the images is to add noise to the input image.
We call this the ``forward timestep matching'' baseline (lines 7c, 8c in \cref{alg:conditional-diffusion}).
In this baseline, $\x_0^\inp$ is transported to timestep $t$ via forward diffusion.
The noised image $\x_t^\inp$ is then compared with $\x_t$.
We now derive a more principled method for guided conditional diffusion,
% based on theoretical analysis
and show that the correct way to match image timesteps is in the \textit{reverse} direction.

\section{Generalized similarity-conditioned guided diffusion and a principled approximation}
\label{sec:proposed-guidance-gradient}

\noindent Conventional classifier guidance is restricted to using a classifier as the conditioning factor.
Furthermore, it also requires training a classifier $p(y \mid \x_t)$ for each intermediate latent state~\cite{song2020score,dhariwal2021diffusion}.
Instead, we extend classifier guidance to the more general setting where the conditioner can be any non-negative function:

\textbf{\textit{Theorem 1:}} When a diffusion model $\epsfunc{t}$ is trained to sample from $q(\x_{0:T})$, the conditional distribution $q(\x_{0:T} \mid \x_0^\inp) \propto q(\x_{0:T})\,\rsim(\x_0, \x_0^\inp)$ can be sampled by
% adding the following guidance gradient in the reverse diffusion step (line 14, \cref{alg:conditional-diffusion})
using the \textit{idealized} guidance gradient $\g_t^*(\x_t)$ during reverse diffusion:
{
\seteqnspacing{4pt}
\begin{align}
    \g_t^*(\x_t)
    &= \nabla_{\x_t} \log\int_{\x_0} q(\x_0 \mid \x_t)\,\rsim(\x_0, \x_0^\inp)\,\dd\x_0\nonumber\\
    &= \nabla_{\x_t} \log \E_{q(\x_0 \mid \x_t)} \Big[ \rsim(\x_0, \x_0^\inp) \Big]
    \label{eqn:guidance-gradient-ideal}
\end{align}
}
\noindent See proof in \cref{app:proof-ideal-g}.
Intuitively, this gradient guides $\x_t$ to increase the expected similarity of its \textit{denoised} version $\x_0 \sim q(\x_0 \mid \x_t)$ with $\x_0^\inp$.
The timesteps of the two images being compared now match, and correspond to \hbox{$t = 0$}, \ie, zero noise.
% The images being compared are now at the same timestep, corresponding to zero noise.
% {\small \textbf{A principled approximation:}}
This makes sense, as we want the eventually denoised image (not the intermediate noisy images) to match the input image.
Unfortunately, the guidance gradient is intractable to compute due to the high-dimensional expectation over $q(\x_0 \mid \x_t)$.
We propose approximating $\E_{q(\x_0 \mid \x_t)} \big[ \rsim(\x_0, \x_0^\inp) \big]$ by a point estimate of $q(\x_0 \mid \x_t)$, namely, its expected value:
% i.e. $\E_{p(\x_0\x_t)} \x_0$. In other words, the guidance gradient would be:
{
\seteqnspacing{4pt}
\begin{align}
    \g_t^*(\x_t) &= \nabla_{\x_t} \log \E_{q(\x_0 \mid \x_t)} \Big[ \rsim(\x_0, \x_0^\inp) \Big]\nonumber\\
    &\approx \nabla_{\x_t} \log \rsim \left( \E_{q(\x_0 \mid \x_t)} \left[ \x_0 \right], \x_0^\inp \right)\nonumber\\
    &\approx \nabla_{\x_t} \log \rsim \left( \mufunc{0}(\x_t)\,,\, \x_0^\inp \right) = \g_t^\weights(\x_t)
    %  \ \text{\small \it(line 7a in \cref{alg:conditional-diffusion})}
    \label{eqn:guidance-gradient-approx}
\end{align}
}
% \begin{align*}
%     \text{Guidance gradient} &= \nabla_{\x_t} \log p(y\mid\x_t)\\
%     &= \nabla_{\x_t}\log\int_{\x_0} p(\x_0 \,\vert\, \x_t)~\underbrace{p(y\mid\x_0)}_{r(\x_0)}~\dd\x_0\\
%     &= \nabla_{\x_t}\log\int_{\x_0} p(\x_0 \,\vert\, \x_t)~r(\x_0)~\dd\x_0\\
%     &= \nabla_{\x_t}\log \E_{p(\x_0\mid\x_t)} r(\x_0)\\
%     &\approx \nabla_{\x_t} \log r \left( \E_{p(\x_0\mid\x_t)} \x_0 \right)\\
%     &= \nabla_{\x_t} \log r \left( \mufunc(\x_t) \right)
% \end{align*}
The point estimate $\E_{q(\x_0\mid\x_t)} \left[ \x_0 \right]$ can be computed analytically by the diffusion model, which we denote by $\mufunc{0}(\x_t)$.
The expression of $\mufunc{0}(\x_t)$ in terms of $\epsfunc{t}(\x_t)$ is shown in line 7a of \cref{alg:conditional-diffusion}.
% and derived in detail in \cref{app:proof-mu-zero}.
Intuitively, since $\epsfunc{t}(\x_t)$ is trained to predict the expected noise $\bepsilon$ that produced $\x_t$ from $\x_0$, the expected value of $\x_0$ given $\x_t$ can be computed by subtracting (a scaled version of) $\epsfunc{t}(\x_t)$ from $\x_t$.
Therefore, our proposed guidance gradient transforms $\x_t$ in the \textit{reverse} direction to \hbox{$t=0$}, matching $\x_0^\inp$.

 Another way to see why our approximation is principled is to examine its value when $t \approx T$.
In forward diffusion, $\x_0$ and $\x_T$ are independently distributed~\cite{sohl2015deep,song2020score}.
This implies that at the beginning of reverse diffusion, the ideal guidance gradient $\g_T^*(\x_T) = \mathbf{0}$ for all $\x_T$ (since the expectation in \cref{eqn:guidance-gradient-ideal} is constant with respect to $\x_T$, therefore the gradient is zero).
This property holds for our approximation $\g_T^\weights(\x_T)$.
Assuming that the diffusion model is well-trained, $\mufunc{0}(\x_T) \approx \E_{q(\x_0 \mid \x_T)} \left[\x_0\right]$ is a constant function of $\x_T$ due to the independence of $\x_0$ and $\x_T$.
Therefore \hbox{$\g_T^\weights(\x_T) = \nabla_{\x_T} \log \rsim(\mufunc{0}(\x_T), \x_0^\inp) \approx \mathbf{0}$}.
However, this property is violated by both baseline guidance gradients $\g_T(\x_T)$ (lines 8b, 8c of \cref{alg:conditional-diffusion}), which, in general, can be non-zero.

% Furthermore, our guidance gradient  $\g_t^\weights(\x_t)$ \textit{leverages the learned diffusion model} to compute the gradient.
As far as we are aware, ours is the first work to \textit{leverage the learned diffusion model $\epsfunc{t}(\x_t)$ to compute the guidance gradient $\g_t^\weights(\x_t)$}.
Furthermore, our method backprops through the network $\epsfunc{t}(\x_t)$ at test time in order to compute $\g_t^\weights(\x_t)$.
In contrast, the baseline guidance gradients $\g_t(\x_t)$
% (lines 8b and 8c of \cref{alg:conditional-diffusion})
are purely analytical and do not use the diffusion model.
We show in \cref{sec:experiments} that utilizing the diffusion model to compute the guidance gradient improves anomaly detection performance.

\section{Extracting anomaly segments from diffusion-edited image using VLP models}

% The diffusion model edited the input image $\x_0^\inp$ by removing anomalies and bringing $\x_0^\inp$ closer to the training distribution, thereby synthesizing $\x_0^\edit$
The diffusion model allows us to synthesize an edited image $\x_0^\edit$ by replacing the anomalous regions of $\x_0^\inp$ with content 
% more closely aligned
similar to the training distribution.
Now, we wish to \textit{analyze} the difference between the two images to detect the modified anomaly segments.
Our pipeline is shown in \cref{fig:pipeline}.
We found that directly comparing pixel intensities between the two images produces very noisy anomaly masks (shown in the extreme right of \cref{fig:pipeline}).
Although the most salient edits made by the diffusion model are the removal of anomaly regions, the model also makes subtle changes across the entire image.
%  that get highlighted by a pixel-wise difference.
Instead, we propose comparing the two images in a feature space that is invariant to subtle intensity changes.
We wish to capture higher-level \textit{semantic} changes, such as the replacement of the brown vehicle with mud in \cref{fig:pipeline}, even though they have similar RGB values.

Our pipeline leverages pre-trained vision-language models.
First, we use MaskCLIP~\cite{dong2023maskclip} to compute CLIP features~\cite{radford2021clip} for each image.
Because MaskCLIP has a large receptive field, it downsamples an input image of size 224$\times$224 to 14$\times$14.
% the computed features are at a lower resolution than the image.
Then, we use FeatUp~\cite{fu2024featup} to upsample the CLIP features back to the full image resolution.
FeatUp is designed to upsample feature embeddings to align them with object boundaries~\cite{fu2024featup}.
Finally, we compute an anomaly score for each pixel as the cosine distance between corresponding CLIP feature vectors.
While this distance-based mask is able to highlight anomaly segments well, we further refine it using the SegmentAnything Model (SAM)~\cite{kirillov2023segment}.
SAM produces a set of accurate, open-world segments per image;
we simply assign to each SAM segment the average cosine distance score averaged across all pixels in that segment.
This approach combines SAM's fine segmentation capability with our coarser difference-based anomaly identification.

{
\newcommand{\emphasis}[1]{#1}
\newcommand{\hthickness}{1.3pt}
\newcommand{\vthickness}{1.3pt}
\newcommand{\NA}{\it \texttt{Not Applicable} \cellcolor{gray!20}}
\newcolumntype{I}{!{\vrule width \vthickness}}
\newcolumntype{|}{!{\vrule width 0.8pt}}
\newcolumntype{:}{!{\vrule width 0pt}}

\newcommand{\redcolor}{red!10}
\newcommand{\greencolor}{green!10}
\newcommand{\bluecolor}{cyan!10}

\newcommand\crule[3][black]{\textcolor{#1}{\rule{#2}{#3}}}
\newcommand{\redbox}{\fcolorbox{black}{\redcolor}{\rule{0pt}{3pt}\rule{3pt}{0pt}}\ }
\newcommand{\greenbox}{\fcolorbox{black}{\greencolor}{\rule{0pt}{3pt}\rule{3pt}{0pt}}\ }
\newcommand{\bluebox}{\fcolorbox{black}{\bluecolor}{\rule{0pt}{3pt}\rule{3pt}{0pt}}\ }

% Vertical cell margins
\renewcommand{\arraystretch}{1.13}

\begin{table*}[th!]

    % Horizontal cell margins
    \setlength\tabcolsep{6pt} % default value: 6pt
    % \small
    \centerline
    {
    \begin{tabular}{Ic|c|cIc|cIc|cI}
        \Xhline{\hthickness}
        % \multicolumn{1}{IcI} & \multicolumn{8}{cI}{Evaluating the overall uncertainty} &  \multicolumn{6}{cI}{Evaluating the density estimator} & \multirow{3}{*}{\makecell{Time\\(ms)}}\\
        % \cline{2-15}
        \multicolumn{3}{IcI}{} & \multicolumn{2}{cI}{RUGD dataset~\cite{RUGD}} & \multicolumn{2}{c|}{RELLIS dataset~\cite{RELLIS}}\\
        \Xcline{4-7}{\hthickness}
        % \multicolumn{2}{c|}{\makecell{AUC-PR\\score vs. anomaly}} & \multicolumn{2}{cI}
        \multicolumn{3}{IcI}{} & {\makecell{AUC-PR} ($\uparrow$)} & 
        {\fonestar~score ($\uparrow$) \vphantom{$\Big\vert$}} & {\makecell{AUC-PR} ($\uparrow$)} &
        {\fonestar~score ($\uparrow$)}\\
            \Xhline{\hthickness}
            \rowcolor{\redcolor}
            & 1 & SSIM~\cite{wang2004ssim} & 0.293 & 0.410 & 0.278 & 0.506\\
            \cline{2-7}
            \rowcolor{\redcolor}
            & 2 & GMM w/ MaskCLIP {\cite{dong2023maskclip} + SAM} &
                {0.554} & {0.587} & {0.549} & {0.534}\\
            \cline{2-7}
            \rowcolor{\redcolor}
            & 3 & Nearest neighbor search {w/ MaskCLIP~\cite{oquab2023dinov2} + SAM} &
                {0.705} & {0.629} & {0.536} & {0.498} \\
            \cline{2-7}
            \rowcolor{\redcolor}
             & 4 & Normalizing flow~\cite{ancha2024icra} w/ {MaskCLIP~\cite{dong2023maskclip} + SAM} &
                {0.616} & {0.596} & {0.541} & {0.545}\\
            \cline{2-7}
            \rowcolor{\redcolor}
            & 5 & Guided diffusion without timestep matching~\cite{sohl2015deep} &
                {0.629} & {0.591} & {0.462} & {0.522} \\
            \cline{2-7}
            \rowcolor{\redcolor}
            \multirow{-6}{*}{\cellcolor{\redcolor}\rotatebox[origin=c]{90}{\it Baselines}} & 6 & Guided diffusion with forward timestep matching &
                {0.697} & {0.578} & {0.462} & {0.514} \\
            \Xhline{\hthickness}
            \rowcolor{\bluecolor}
            & 7 & Ours \textit{with} ResNet~\cite{he2016resnet} &
                {0.424} & {0.524} & {0.255} & {0.484}\\
            \cline{2-7}
            \rowcolor{\bluecolor}
            & 8 & Ours \textit{without} SAM~\cite{kirillov2023segment} &
                {0.645} & {0.737} & {0.396} & {{0.546}} \\
            \cline{2-7}
            \rowcolor{\bluecolor}
            & 9 & Ours \textit{with} {RGB features} &
                {0.302} & {0.591} & {0.144} & {0.482}\\
            \cline{2-7}
            \rowcolor{\bluecolor}
            & 10 & Ours \textit{with} {DINOv2~\cite{oquab2023dinov2}} &
                {0.685} & {0.598} & {0.416} & {0.518}\\
            \cline{2-7}
            \rowcolor{\bluecolor}
            & 11 & Ours \textit{with} {ViT~\cite{alexey2021vit}} &
                {0.692} & {0.571} & {0.255} & {0.493} \\
            \cline{2-7}
            \rowcolor{\bluecolor}
            & 12 & Ours \textit{with} {CLIP~\cite{radford2021clip}} &
                {0.694} & {0.585} & {0.499} & {0.532} \\
            \cline{2-7}
            \rowcolor{\bluecolor}
            \multirow{-7}{*}{\rotatebox[origin=c]{90}{\it Ablations}} & 13 & Ours \textit{with} {DINOv1~\cite{caron2021dino}} &
            {0.665} & {0.622} & {0.416} & {0.521}\\
            \Xhline{\hthickness}
            \rowcolor{\greencolor}
            & 14 & \textbf{Ours \textit{without} FeatUp} &
                {\bf 0.724} & {\bf 0.858} & {0.475} & \textbf{0.549}\\
            \cline{2-7}
            % \makecell{\textbf{Ours} \fontsize{6}{6}\selectfont (Reverse matching +\\\fontsize{6}{6}\selectfont MaskCLIP + SAM + FeatUp)
            \rowcolor{\greencolor}
            \multirow{-2}{*}{\rotatebox[origin=c]{90}{\it Ours}} & 15 & \textbf{Ours} {\fontsize{7}{7}\selectfont (Reverse matching + MaskCLIP + SAM)} {\bf \textit{with} FeatUp}
            & {0.709} & {0.599} & {\textbf{0.568}} & {0.540} \\
            \Xhline{\hthickness}
        \end{tabular}
        }
    \vspace{0.4em}
    \caption{
        Anomaly detection accuracy on RUGD~\cite{RUGD} dataset and the RELLIS-3D~\cite{RELLIS} dataset.
        \greenbox Our methods (in green) are compared against \redbox baselines (in red), and \bluebox ablations (in blue) removing or changing one component of our method at a time. See text for details.
        % \redbox Baselines are in red, \bluebox ablations are in blue and our \greenbox method is in green.
        \vspace{-1.2em}
    }
    \label{table:main-results-rugd}
    \vspace*{-1em}
\end{table*}
}
\section{Experiments}
\label{sec:experiments}

\subsection{Dataset and experimental design}
% We validate our approach using the following datasets of real-world camera images collected in off-road environments using mobile robot platforms and manually labeled by humans for semantic segmentation:\\

We wish to identify pixels belonging to OOD anomaly objects.
From the full semantic label set $\classes = \{1, \dots, C\}$, we define a subset $\classesID \subset \classes$ as the set of in-distribution classes, and the remaining subset $\classesOOD = \classes \setminus \classesID$ as out-of-distribution classes. 
We split the overall dataset such that
pixels in the training dataset $\dataset$ belong exclusively to $\classesID$, whereas test images contain pixels drawn from both $\classesID$ and $\classesOOD$.
% We evaluate uncertainty estimation performance on a separate dataset that additionally contain OOD obstacle segments, such as buildings and vehicles that are not present in the training images.
Importantly, we assume that the learner has no access to any amount of real or synthetic OOD data at training time, and we do not make any assumptions about the nature of OOD data that the model can expect at test time.

We quantitatively validate our approach using two off-road land navigation datasets: the RUGD~\cite{RUGD} dataset and the RELLIS-3D~\cite{RELLIS} dataset.
Both are real-world datasets containing camera images collected in off-road environments using mobile robot platforms with manually labeled pixel-wise class annotations.
They contain 7,453 and 6,235 labeled images, respectively.
% from 17 and 5 scenes.
We split the semantic categories into in-distribution labels that contains mostly natural features and vegetation such as \texttt{dirt}, \texttt{grass}, \texttt{sky} etc., whereas classes like \texttt{vehicle}, \texttt{building}, \texttt{person} are defined as out-of-distribution.
See \cref{app:rugd-dataset} for a full list of in-distribution and out-of-distribution classes for each dataset.

\subsection{Evaluation metrics}

We evaluate our approach using the following metrics. The results of our evaluation are presented in \cref{table:main-results-rugd}.
\textBF{(i) AUC-PR:}
We compute the area under the precision-recall curve between the per-pixel anomaly score, and the ground-truth classification of the pixel as in-distribution.
% (highlighted in blue in Table~\ref{table:main-results-rugd}).
\textBF{(ii) \fonestar~score:} We borrow this metric from existing anomaly detection benchmarks for structured urban-driving scenes~\cite{chan2021smiyc}.
The \fonestar-score summarizes true positive, false positive and false negative detections averaged over different detection thresholds, and normalized by the size of ground- truth segments to prevent large objects dominating the metric.
Not all classes in $\classesID$ are equally in-distribution; some classes (e.g. \texttt{grass}) occur more frequently than others (e.g. \texttt{rock-bed}).
Therefore, we weight the classes in $\classesID$ by their frequency in the training dataset when computing AUC-PR and \fonestar~scores.

% \noindent\textbf{(iii) \small FPR@x\%TPR:} Also used by \citet{chan2021smiyc}, this computes the false positive rate of anomaly detections at a desired true positive rate.
% We report this score for a varying number of TPR thresholds.

\subsection{Baselines and ablations}
We compare against multiple baselines in Table~\ref{table:main-results-rugd}.
\smalltextbf{SSIM} (row 1): We use the structural similarity index (SSIM)~\cite{wang2004ssim} as an anomaly score that compares images using pixel intensities (and not a learned feature space).
Baselines 2-4 represent the conventional paradigm of directly predicting anomaly scores from pixels. We fit a 
\smalltextbf{GMM} (row 2) with 20 components on pixel-level MaskCLIP~\cite{dong2023maskclip} features and uses the negative log-likelihood as the anomaly score.
\smalltextbf{Nearest-neighbor search} (row 3): is akin to memorizing the training dataset. We collect 50,000 MaskCLIP pixel feature vectors sampled from training images. For a given feature vector of pixel $i$ at test time, we output the distance of the nearest neighbor in the collected set. This is treated as the anomaly score.
\smalltextbf{Normalizing flow}~\cite{ancha2024icra,charpentier2022natural} (row 4): We train a normalizing flow
% which is among the best performing generative models prior to diffusion models,
on MaskCLIP pixel features from the training set similar to the GMM baseline. We follow \citet{ancha2024icra} and use a GMM as a stronger base distribution. Negative log-likelihoods of the normalizing flows, which can be computed exactly~\cite{papamakarios2021normalizing}, are used as anomaly scores.
\smalltextbf{Diffusion guidance baselines}: We compare against vanilla guided diffusion with no timestep matching~\cite{sohl2015deep} (line 8b of \autoref{alg:conditional-diffusion}), and forward timestep matching (lines 7c, 8c of \autoref{alg:conditional-diffusion}).
% \textbf{Autoencoder-only} (row 1): We compare against the anomaly detection method of \citet{richter2017safe} by using the patch-based autoencoder to predict pixel-wise reconstruction scores. The scores are treated analogous to the (negative of)  density.
We also perform ablation experiments in rows 7-14, by either removing one of our contributions at a time, or changing the VLP feature space in the analysis component.
% The key takeaways from these experiments are as follows.
We find that our method outperforms all baselines and ablations, on both datasets and evaluation metrics.
On RUGD, we find that not using FeatUp~\cite{fu2024featup} can improve quantitative performance.
% at the cost of reducing resolution.
We hypothesize that this might be because FeatUp can average upsampled pixel features near segment boundaries,
% affecting the comparison between both images.
affecting the resulting cosine distances.
However, FeatUp can be useful for certain applications where high-resolution is crucial when detecting anomalies.
We use FeatUp when generating all visualizations in this paper.

We also apply our method to an under-water navigation dataset~\cite{manderson2020rss} where the training distribution contains images of the ocean bed, corals and plants.
We are unable to perform a quantitative evaluation due to the lack of ground truth semantic segmentation labels.
We qualitatively evaluate our method on anomalous images that contain fish, divers and robots in \cref{fig:pull-figure}.
We observe that the diffusion model makes interesting edits to the input image.
It either removes anomalies altogether, or blends them into the background.
Both types of edits are detected by our analysis pipeline.
% See our \href{\website}{project website} and appendix for more qualitative results.

\smalltextbf{Computational considerations:}
results in this paper use DDPM, which is computationally intensive ($\approx$18s per image).
While the focus of this work is on developing a performant analysis-by-synthesis framework for anomaly detection,
we can significantly improve efficiency using accelerated sampling techniques like DDIM~\cite{songdenoising} ($\approx$1.8s per image) with only a slight loss in accuracy.
%  see \cref{app:ddim} for details.

\vspace{-0.3\baselineskip}

\section{Related work}

\todo[inline]{
    \begin{itemize}[leftmargin=*]
        \item Talk about classifier-free guidance~\cite{} and why it's not applicable in our settings -- we would need to train a conditional generative model, which in our case would mean conditioning on OOD images that we assume we don't have access to.
    \end{itemize}
}

Prior work on anomaly detection~\cite{anomaly_survey} has predominantly focused on \textit{discriminative} models that directly map pixels to anomaly scores, classifying features at the pixel- or segment- level as in-distribution or anomalies~\cite{ancha2024icra,ulmer2021survey,charpentier2020posterior,charpentier2022natural,grcic2024dense,liu2023residual,hendrycks2016baseline,lakshminarayanan2017simple,lee2018simple,gudovskiy2023concurrent,tian2022pixel,bevandic2022dense,chan2021entropy,nayal2023rba,ackermann2023maskomaly} using principal component analysis (PCA) \cite{candes2011robust, scholkopf1997kernel, zhou2017anomaly}, random feature projection~\cite{li2006very, pang2018learning, pevny2016loda}, novelty functions \cite{pimentel2014review,richter2017autonomous}, feature-space comparison of neural embeddings \cite{erfani2016high, ionescu2019object, xu2015learning, yu2018netwalk} and more recently, evidential uncertainty estimation~\cite{charpentier2020posterior,charpentier2022natural,amini2020deep} for off-road navigation~\cite{ancha2024icra,cai2024evora,cai2024pietra,sirohi2023uncertainty}.
% These approaches predict anomalies directly from the input image.
Other works that take an analysis-by-synthesis~\cite{bever2010analysis,yuille2006vision} approach
% to anomaly detection
use \textit{generative} models such as autoencoders~\cite{inpainting,dan2015learning} and GANs~\cite{gan_radiographs, anogan, fast-anogan, ganomaly, skip-ganomaly}.
However, these approaches can suffer from high false-positive rates~\cite{anomaly_survey,inpainting}.
More recently, a small set of works have employed diffusion models for anomaly detection~\cite{wolleb2022diffusion,pinaya2022fast,zhang2023diffusionad,mousakhan2023anomaly}.
However, these works focus on specialized domains like medical imaging~\cite{wolleb2022diffusion,pinaya2022fast} and industrial inspection~\cite{zhang2023diffusionad,mousakhan2023anomaly}; to the best of our knowledge, we are the first to use this approach for natural images in off-road navigation.
Furthermore, \citet{zhang2023diffusionad} train on synthetic anomaly examples, whereas our method does not require real or synthetic OOD examples during training.

\section{Conclusions}

\noindent In this work, we presented an \textit{analysis-by-synthesis} approach for pixel-wise anomaly detection in off-road images.
Given an input image, we used a diffusion model to \textit{synthesize} an edited image that removes anomalies while keeping the remaining image unchanged.
% closer to the training distribution by removing anomalies.
We then formulated anomaly detection as \textit{analyzing} which image segments were modified by the diffusion model.
% We propose a novel approximate inference technique based on a principled analysis of guided diffusion that bootstraps the diffusion model to compute guidance gradients.
We proposed a novel inference approach for guided diffusion by theoretically analyzing the ideal guidance gradient and deriving a principled approximation.
Unlike prior methods, this approach bootstraps the diffusion model to predict edits to the image.
% Our editing technique is purely test-time and does not modify how diffusion models are trained.
Our editing technique is purely test-time and can be integrated into existing workflows without re-training or finetuning.
% Furthermore, our method is interpretable --- by synthesizing images that remove anomalies, we can visualize why the model believes certain regions are OOD.
Finally, we presented a combination of vision-language foundation models to compare pixels in a more semantically meaningful feature space in order to identify segments that were modified by the diffusion model.
We hope this work paves the way towards generative approaches for accurate and interpretable anomaly detection for off-road navigation.

% \input{sections/notes.tex}

%%%%%%%%%%%%%%%%%%%%%%%%%%%%%%%%%%%%%%%%%%%%%%%%%%%%%%%%%%%%%%%%%%%%%%%%%%%%%%%%


%%%%%%%%%%%%%%%%%%%%%%%%%%%%%%%%%%%%%%%%%%%%%%%%%%%%%%%%%%%%%%%%%%%%%%%%%%%%%%%%

% \clearpage
% \newpage
\bibliographystyle{unsrtnat}
\bibliography{main}

%%%%%%%%%%%%%%%%%%%%%%%%%%%%%%%%%%%%%%%%%%%%%%%%%%%%%%%%%%%%%%%%%%%%%%%%%%%%%%%%

%%%%%%%%%%%%%%%%%%%%%%%%%%%%%%%%%%%%%%%%%%%%%%%%%%%%%%%%%%%%%%%%%%%%%%%%%%%%%%%%

\clearpage

\begin{appendices}
  
  % Appendix title
  \begingroup
  \centering
  % \twocolumn[
  \oldtwocolumn[  % after redefining \twocolumn to insert image before abstract
    \centering
    {\fontsize{25pt}{25pt}\selectfont Appendix}\\
    \vspace{1.5em}
    {\LARGE \bf \textit{Anomalies-by-Synthesis}: Anomaly Detection using\\\vspace{4pt}Generative Diffusion Models for Off-Road Navigation}

    \vspace{2em}
    \centering{
      \fontsize{11pt}{11pt}\selectfont
      \authornames\\
    }
    \vspace{2em}
    {\small Website: \href{\website}{\color{deeppink}\texttt{\website}}$^\dagger$}

    \vspace{4em}
    \centerline{
      \includegraphics[width=0.98\textwidth]{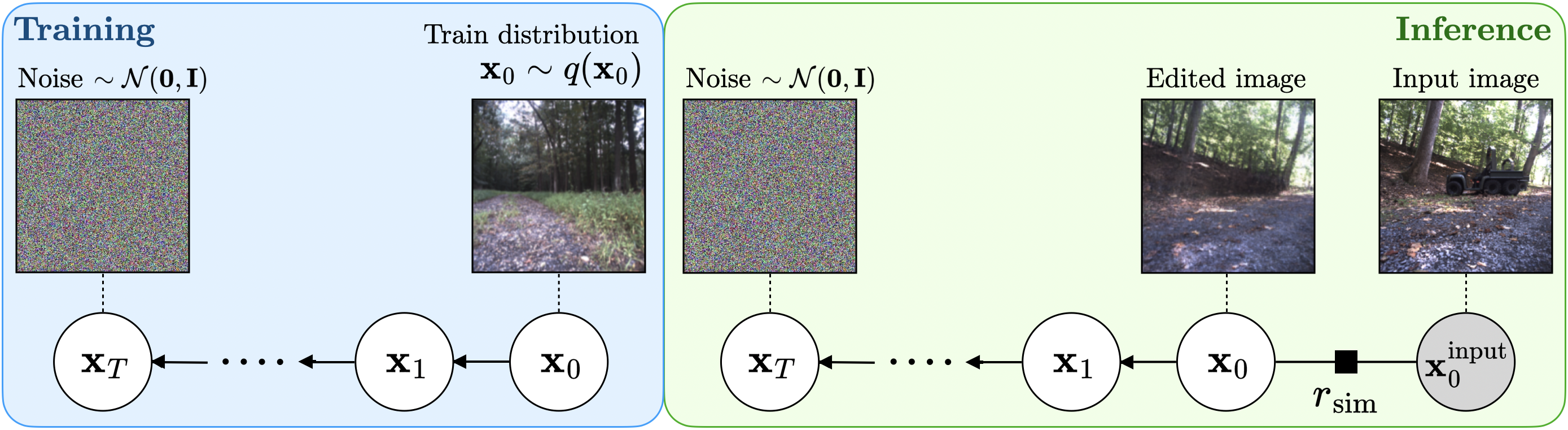}
    }
\captionof{figure}
{
    \textbf{Probabilistic graphical model of the \textit{conditional} forward diffusion process.}
    We wish to sample the random variable $\x_0$ corresponding to the training data distribution $q(\x_0)$.
    The \textit{directed} edges between $\x_{t-1}$ and $\x_t$ (for $t = 1, \dots, T$) correspond to the vanilla forward diffusion process.
    Each directed edge denotes the sampling distribution $q(\x_t \mid \x_{t-1})$ which successively adds a small amounts of Gaussian noise: $q(\x_t \mid \x_{t}) = \N\left( \x_t\,;\,\sqrt{1-\beta_t}\,\x_{t-1}, \beta_t \bf{I} \right)$~\cite{sohl2015deep,ho2020denoising}.
    However, we are interested in sampling from $q(\x_0 \mid \x_0^\inp) = q(\x_0),\rsim(\x_0, \x_0^\inp)$.
    This objective corresponds to adding an additional \textit{undirected factor} $\rsim(\x_0, \x_0^\inp)$ between $\x_0$ and $\x_0^\inp$; $\x_0^\inp$ is treated as constant.
    Our task is to perform inference over this graphical model and sample from $q(\x_0 \mid \x_0^\inp)$ using a diffusion model that was trained to perform reverse diffusion in the absence of the $\rsim$ factor. 
}

    \vspace{2em}
  ]
  \endgroup

  \section{Proof: Generalized conditional guidance gradient}
\label{app:proof-ideal-g}

Classifier guidance is not only restricted to classifiers, it also requires training a classifier $p(y \mid \x_t)$ for each intermediate latent state~\cite{song2020score,dhariwal2021diffusion}.
First, we extend classifier guidance to the more general setting where the conditioner is any non-negative function:

\textbf{\textit{Theorem 1:}} When a diffusion model $\epsfunc{t}$ is trained to sample from $q(\x_0)$, the conditional distribution $q(\x_0 \mid \x_0^\inp) \propto q(\x_0)\,\rsim(\x_0, \x_0^\inp)$ can be sampled by
% adding the following guidance gradient in the reverse diffusion step (line 14, \cref{alg:conditional-diffusion})
using the following guidance gradient during reverse diffusion: 
% evaluated at $\x_t = \mufunc{t+1}(\x_{t+1})$:
\begin{align}
    \g_t(\x_t)
    &= \nabla_{\x_t} \log\int_{\x_0} q(\x_0 \mid \x_t)\,\rsim(\x_0, \x_0^\inp)\,\dd\x_0\nonumber\\
    &= \nabla_{\x_t} \log \E_{q(\x_0 \mid \x_t)} \Big[ \rsim(\x_0, \x_0^\inp) \Big]
\end{align}

\subsection{Proof using the variational inference perspective}

\newcommand{\blueunderbrace}[1]{{\color{blue}\underbrace{\color{black}#1}_{}}}

\begin{align*}
    &q(\x_t \mid \x_{t+1}, \x_0^\inp)\propto q(\x_t, \x_{t+1} \mid \x_0^\inp)\\
    &= \int_{\x_0} q(\x_t, \x_{t+1}, \x_0 \mid \x_0^\inp)\,\dd\x_0\\
    &= \int_{\x_0} q(\x_0 \mid \x_0^\inp)\,q(\x_t, \x_{t+1} \mid \x_0, \x_0^\inp) \,\dd\x_0\\
    &= \int_{\x_0} q(\x_0 \mid \x_0^\inp)\,\blueunderbrace{q(\x_t, \x_{t+1} \mid \x_0)}\,\dd\x_0\\[-9pt]
    &\hphantom{=\ \ }\text{\color{blue}\small\it(since $\x_{1:T}$ is independent of $\x_0^\inp$ conditioned on $\x_0$)}\\
    % &= \int_{\x_0} q(\x_0)\,q(\x_t, \x_{t+1} \mid \x_0) \,\dd\x_0 \ \ \text{\color{blue}\small(forward diffusion process is the same in both cases)}\\
    &\propto \int_{\x_0} \blueunderbrace{q(\x_0)\,\rsim(\x_0, \x_0^\inp)}\,q(\x_t, \x_{t+1} \mid \x_0) \,\dd\x_0\\[-9pt]
    &\hphantom{=\ \ }\text{\color{blue}\small\it(by the definition of $q(\x_0 \mid \x_0^\inp) \propto q(\x_0)\,\rsim(\x_0, \x_0^\inp)$)}\\
    &= \int_{\x_0} \rsim(\x_0, \x_0^\inp)\,q(\x_t, \x_{t+1}, \x_0) \,\dd\x_0\\
    &= \int_{\x_0} \rsim(\x_0, \x_0^\inp)\,q(\x_{t+1})\,q(\x_t \mid \x_{t+1})\,q(\x_0 \mid \x_t, \x_{t+1}) \,\dd\x_0\\
    &= \int_{\x_0} \rsim(\x_0, \x_0^\inp)\,q(\x_{t+1})\,q(\x_t \mid \x_{t+1})\,\blueunderbrace{q(\x_0 \mid \x_t)} \,\dd\x_0\\[-9pt]
    &\hphantom{=}\hspace{3em}\text{\color{blue}\small\it(since $\x_0$ and $\x_{t+1}$ are independent conditioned on $\x_t$)}\\
    &\propto q(\x_t \mid \x_{t+1})\int_{\x_0} q(\x_0 \mid \x_t) \,\rsim(\x_0, \x_0^\inp)\,\dd\x_0\\
    &= q(\x_t \mid \x_{t+1}) \,\E_{q(\x_0 \mid \x_t)} \left[\rsim(\x_0, \x_0^\inp)\right]
\end{align*}

Therefore, following the same analysis as the first-order Gaussian approximation, the guidance gradient is $\nabla_{\x_t} \log \E_{q(\x_0 \mid \x_t)} \big[\rsim(\x_0, \x_0^\inp)\big]$ evaluated at \hbox{$\x_t = \boldsymbol\mu_{t+1}(\x_{t+1})$}.

\subsection{Proof using the score functions perspective}
The score function of intermediate states $\x_t$ of the vanilla forward diffusion process is defined as \hbox{$\s(\x_t) = \nabla_{\x_t} \log q(\x_t)$}.
However, we're interested in the score function of the \textit{conditional} forward diffusion process \hbox{$\s(\x_t \mid \x_0^\inp) = \nabla_{\x_t} \log q(\x_t \mid \x_0^\inp)$}.
The additional term that needs to be added to $\s(\x_t)$ to obtain $\s(\x_t \mid \x_0^\inp)$ is the guidance gradient.
Therefore, we now derive $\s(\x_t \mid \x_0^\inp)$ in terms of $\s(\x_t)$: 
\newcommand{\hshift}{-6.5em}
\begin{align*}
    \underbrace{\s(\x_t \mid \x_0^\inp)}_{\text{conditional score function}} &= \nabla_{\x_t}\log q(\x_t \mid \x_0^\inp)\\
    &\hspace{\hshift}= \nabla_{\x_t} \log \int_{\x_0} q(\x_t, \x_0 \mid \x_0^\inp) \,\dd\x_0\\
    &\hspace{\hshift}= \nabla_{\x_t} \log \int_{\x_0} q(\x_0 \mid \x_0^\inp)\,q(\x_t \mid \x_0, \x_0^\inp) \,\dd\x_0\\
    &\hspace{\hshift}= \nabla_{\x_t} \log \int_{\x_0} q(\x_0 \mid \x_0^\inp)\,\blueunderbrace{q(\x_t \mid \x_0)} \,\dd\x_0\\[-9pt]
    &\hspace{\hshift}\hspace{1em}\text{\color{blue}\small\it(since $\x_{1:T}$ is independent of $\x_0^\inp$ conditioned on $\x_0$)}\\
    &\hspace{\hshift}= \nabla_{\x_t} \log \int_{\x_0} \blueunderbrace{q(\x_0)\,\rsim(\x_0, \x_0^\inp)}\,q(\x_t \mid \x_0) \,\dd\x_0\\[-9pt]
    &\hspace{\hshift}\hphantom{=\ \ }\text{\color{blue}\small\it(by the definition of $q(\x_0 \mid \x_0^\inp) \propto q(\x_0)\,\rsim(\x_0, \x_0^\inp)$)}\\
    &\hspace{\hshift}= \nabla_{\x_t} \log \int_{\x_0} \rsim(\x_0, \x_0^\inp)\,q(\x_t , \x_0) \,\dd\x_0\\
    &\hspace{\hshift}= \nabla_{\x_t} \log \int_{\x_0} \rsim(\x_0, \x_0^\inp)\,q(\x_t)\,q(\x_0 \mid \x_t) \,\dd\x_0\\
    &\hspace{\hshift}= \nabla_{\x_t} \log q(\x_t) + \nabla_{\x_t} \log\int_{\x_0} \hspace{-0.7em}q(\x_0 \mid \x_t)\,\rsim(\x_0, \x_0^\inp)\,\dd\x_0\\
    &\hspace{\hshift}= \nabla_{\x_t} \log q(\x_t) + \nabla_{\x_t} \log \E_{q(\x_0 \mid \x_t)} \left[ \rsim(\x_0, \x_0^\inp) \right]\\
    &\hspace{\hshift}= \s(\x_t) + \underbrace{\nabla_{\x_t} \log \E_{q(\x_0 \mid \x_t)} \left[ \rsim(\x_0, \x_0^\inp) \right]}_\text{guidance gradient}
\end{align*}

  \section{Training on the RUGD dataset}
\label{app:rugd-dataset}

\subsection{Information about the RUGD dataset}

\begin{figure*}[h]   
    \centering
    \includegraphics[width=0.89\linewidth]{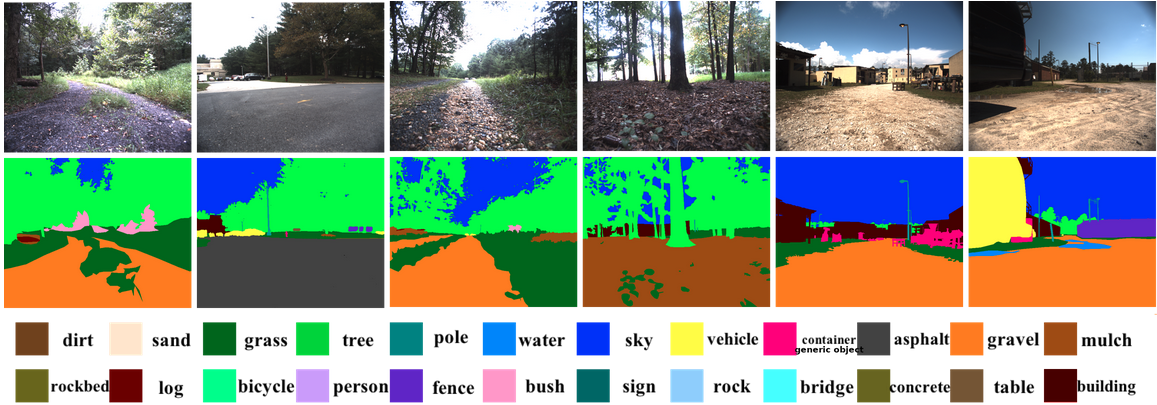}
    \caption[The RUGD dataset]
    {Examples of video frames, annotations and semantic classes from the full RUGD dataset \cite{RUGD}.}
    \label{fig:rugd}
\end{figure*}

The RUGD dataset (\cref{fig:rugd}, \cite{RUGD}) is an off-road dataset of video sequences captured from a small, unmanned mobile robot traversing in unstructured environments. It contains over 7,453 labeled images from 17 scenes, annotated with pixel-level segmentation over 24 semantic classes. The annotated frames are spaced five frames apart.

We split the 24 semantic categories as 16 in-distribution labels: $\classesID$ = {\small\{\texttt{dirt}, \texttt{sand}, \texttt{grass}, \texttt{tree}, \texttt{pole}, \texttt{sky}, \texttt{asphalt}, \texttt{gravel}, \texttt{mulch}, \texttt{rock-bed}, \texttt{log}, \texttt{fence}, \texttt{bush}, \texttt{sign}, \texttt{rock}, \texttt{concrete}\}},
and 8 OOD labels corresponding to ``obstacle'' classes: $\classesOOD$ = {\small\{\texttt{vehicle}, \texttt{container/generic-object}, \texttt{building}, \texttt{bicycle}, \texttt{person}, \texttt{bridge}, \texttt{picnic-table}, \texttt{water}\}}

\subsection{Training a diffusion model on the RUGD Dataset}
\label{app:rugd-dataset-details}

\begin{figure*}[h]   
    \centerline{
        \includegraphics[width=0.532\textwidth]{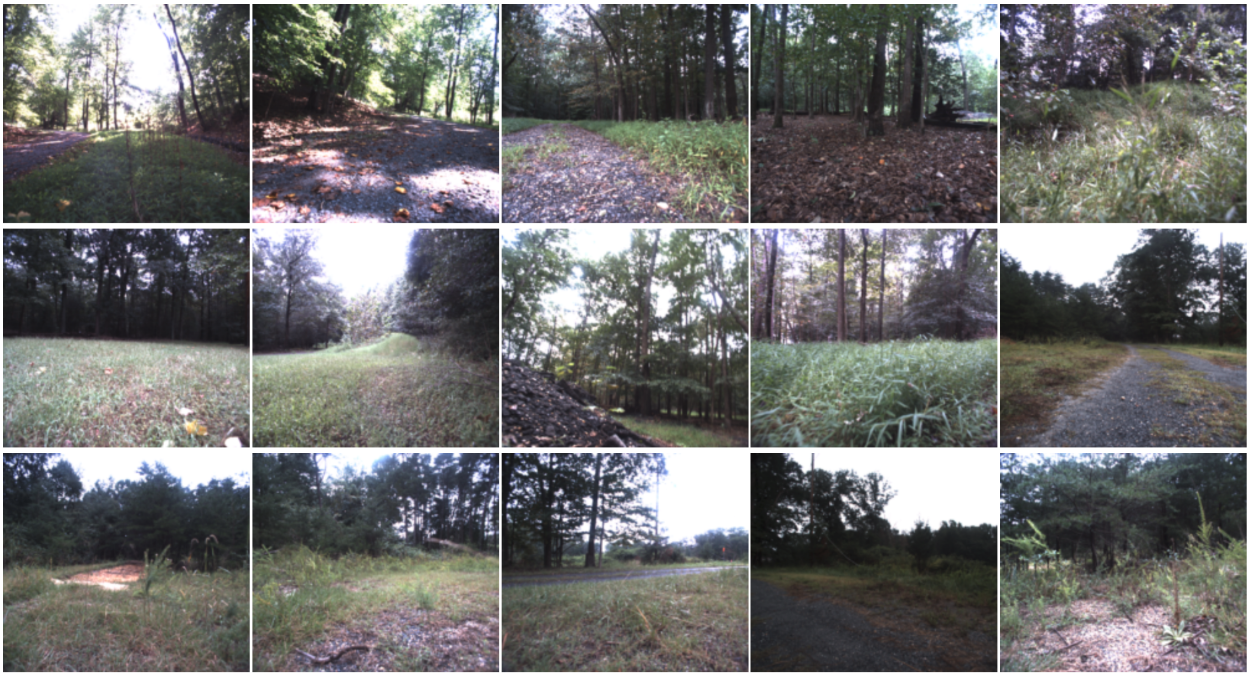}
        \includegraphics[width=0.51\textwidth]{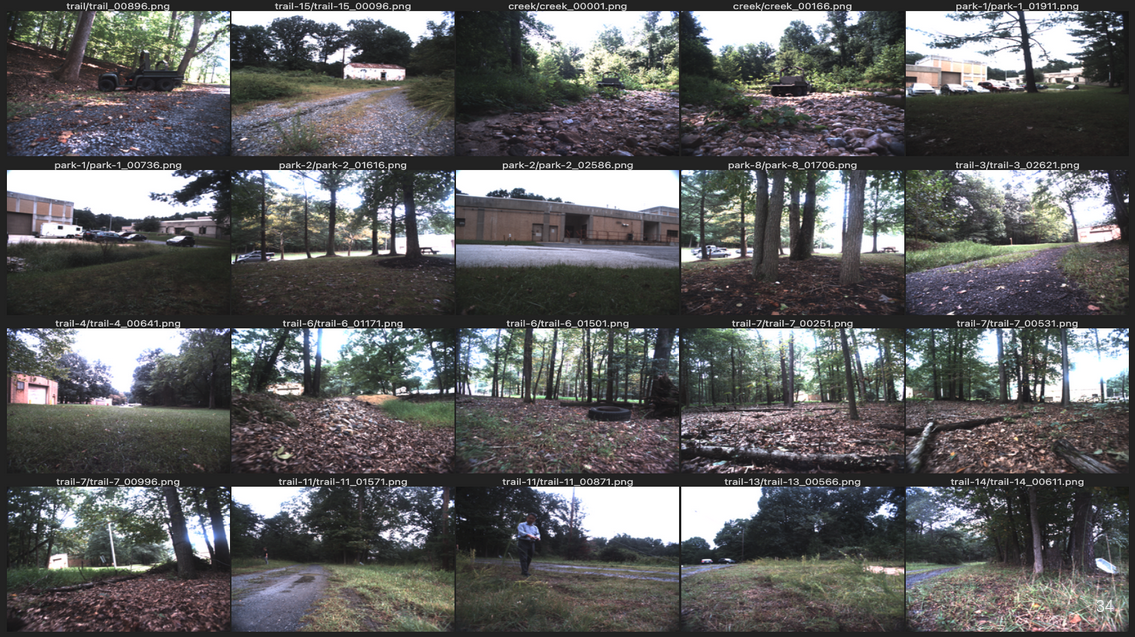}
        % % {
        %     \setlength{\fboxrule}{1pt}
        %     \setlength{\fboxsep}{0pt}
        %     \fbox{\includegraphics[width=0.525\textwidth,trim={4pt 3.5pt 3.5pt 5pt},clip]{diffunc-thesis/figures/rugd_training.png}}
        % % }
        % % {
        %     \setlength{\fboxrule}{0pt}
        %     \setlength{\fboxsep}{0pt}
        %     \fbox{\includegraphics[width=0.51\textwidth,trim={0 0 0 5pt},clip]{diffunc-thesis/figures/rugd_anomalies.png}}
        % % }
    }
    \caption[Training set images for RUGD diffusion model]
    {
        \textbf{In-distribution and out-of-distribution images from the RUGD dataset.}
        \textbf{\textit{Left:}} Examples of the in-distribution images on which our RUGD diffusion model was trained. In general, these images contain natural, off-road vegetation --- a mixture of forest, meadow, mulch, and paths, without any humans or artificial constructions like buildings or vehicles.
        \textbf{\textit{Right:}} Examples of held-out, out-of-distribution RUGD images the robot might encounter. These contains anomaly objects like buildings and vehicles.
        The diffusion model trained on the images on the left must remove anomalies from the images on the right.
    }
    \label{fig:rugd-id-and-ood-images}
\end{figure*}

\begin{figure}
    \centerline{
        \includegraphics[width=1.05\linewidth]{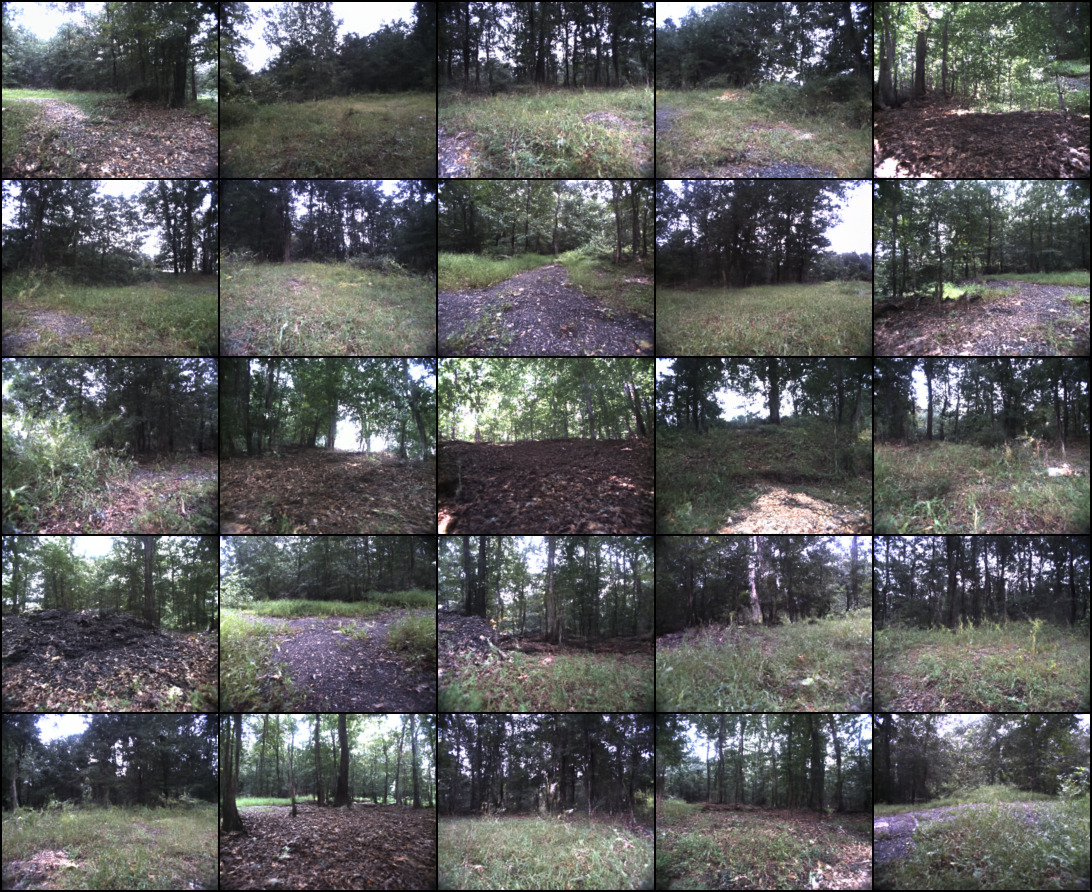}
    }
    \caption{
        Samples generated from the trained diffusion model (without conditioning).
        The training data is shown in \cref{fig:rugd-id-and-ood-images} \textit{(left)}.
        The generated samples are photorealistic, and appear very similar to the training images.
    }
    \label{fig:rugd-diffusion-samples}
\end{figure}

Our diffusion model is trained on samples from the RUGD train split that does not contain  humans and artificial constructs (\cref{fig:rugd-id-and-ood-images} \textit{(left)}). The out-of-distribution and anomalous images are `held out' for evaluation (\cref{fig:rugd-id-and-ood-images} \textit{(right)}). As can be seen from \cref{fig:rugd-diffusion-samples}, our trained diffusion model successfully generates realistic images containing only in-distribution classes such as trees, grass, ground \textit{etc.}

% \subsection{Performance on RUGD Dataset}

% We evaluate our method on the RUGD \cite{RUGD} dataset and share qualitative results in \cref{fig:results}. The diffusion model is trained on RUGD data without artificial constructs. At test-time, we present our method with OOD images containing anomalies from held-out classes.
% % See \cref{sec:off-road} for examples of training and anomalous data.
% \cref{fig:ablations} shows the impact of each component of our \textit{analysis} pipeline.

% \begin{figure}[h]    
%     \centerline{
%         \includegraphics[width=0.95\linewidth]{diffunc-thesis/figures/results.png}
%     }
%     \caption[Results on anomalous samples from RUGD]
%     {Qualitative results on small, anomalous examples from the RUGD dataset.
%     % Our pixel-level uncertainty estimates do generally highlight anomalous regions, but our mask-level estimates is where \diffunc{} truly shines.
%     Our full \diffunc{} pipeline does particularly well at detecting small and camouflaged/human-imperceptible anomalies.}
%     \label{fig:results}
% \end{figure}

% We show the outputs after each stage of our pipeline in \cref{sec:pipeline}, and share the impact of each component on model \diffunc{} via ablations in \cref{app:ablations}.
%We compare with baselines in \cref{sec:baselines}.
% We share performance on other datasets in Section
%\ref{sec:rellis-performance} and
% \ref{sec:urban-performance}.

%\subsection{Approximating the Guidance Likelihood} \label{sec:approx}

%[derivation and justification for the P-sample guidance]

\subsection{Tuning the Guidance Strength}
\label{sec:guidance}

\begin{figure}[h]   
    \centerline{
        \includegraphics[width=1\linewidth]{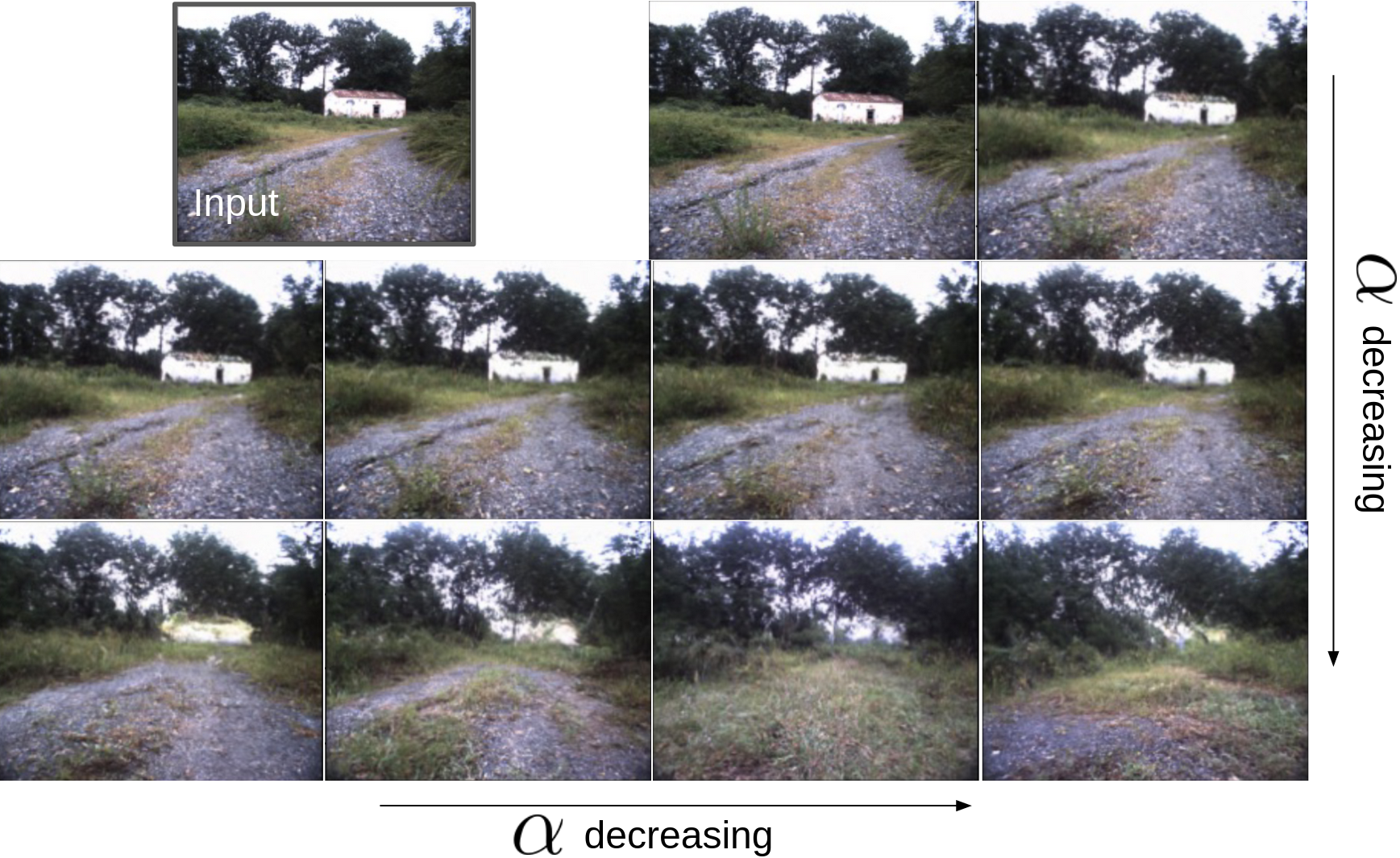}
        % \setlength{\fboxrule}{1pt}
        % \setlength{\fboxsep}{0pt}
        % \fbox{\includegraphics[width=0.53\linewidth,trim={2pt 1.5pt 0pt 0pt}]{diffunc-thesis/figures/limitations.png}}
    }
    \caption
    {
        The strength of the diffusion model's guidance gradient can be tuned by a hyperparameter $\alpha$. As $\alpha$ decreases, the guidance enforcing similarity between the input image and the image synthesized by the model reduces.
    }
    \label{fig:guidance-strength-and-limitations}
\end{figure}

The strength of the guidance term in our diffusion model can be tuned to enforce a variable level of consistency between the input image and the image synthesized by the diffusion model. \cref{fig:guidance-strength-and-limitations} shows the impact of the guidance term on the image generated, for an example input image from the RUGD dataset.

%\subsection{Implementation Details}\label{sec:details}

%[hyperparameters (incl. guidance strength), training schedules, which classes were held out for which experiments, \textit{etc.}]

  % \input{appendix/ddim.tex}
  % \input{appendix/clevr.tex}

  % \input{appendix/instructions.tex}

% \section*{ACKNOWLEDGMENT}

% The preferred spelling of the word ÒacknowledgmentÓ in America is without an ÒeÓ after the ÒgÓ. Avoid the stilted expression, ÒOne of us (R. B. G.) thanks . . .Ó  Instead, try ÒR. B. G. thanksÓ. Put sponsor acknowledgments in the unnumbered footnote on the first page.

\end{appendices}

%%%%%%%%%%%%%%%%%%%%%%%%%%%%%%%%%%%%%%%%%%%%%%%%%%%%%%%%%%%%%%%%%%%%%%%%%%%%%%%%

% References are important to the reader; therefore, each citation must be complete and correct. If at all possible, references should be commonly available publications.

\end{document}